\documentclass[a4paper,12pt,times,numbered,print,index]{Classes/PhDThesisPSnPDF}
\input{Preamble/preamble}

\title{Parameter Optimization using high-dimensional Bayesian Optimization}

\subtitle{Bachelor Thesis}

\author{A. K. David Yenicelik}

\dept{Department of Computer Science}

\university{Swiss Federal Institute of Technology in Zurich}
\crest{\includegraphics[width=0.7\textwidth]{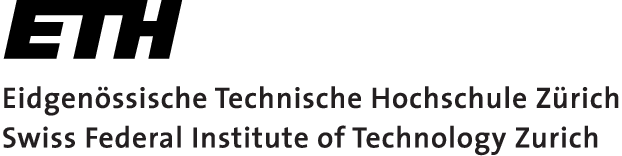}}

\supervisor{Prof. A. Krause \newline
J. Kirschner \newline
M. Mutný }

\degreetitle{Bachelor of Science, Computer Science}

\degreedate{August 2018} 

\subject{LaTeX} 

\ifdefineAbstract
\fi

\ifdefineChapter
\fi

\begin{document}

\frontmatter

\maketitle

\begin{abstract}

In this thesis, I explore the possibilities of conducting Bayesian optimization techniques in high dimensional domains.
Although high dimensional domains can be defined to be between hundreds and thousands of dimensions, we will primarily focus on problem settings that occur between two and 20 dimensions.
As such, we focus on solutions to practical problems, such as tuning the parameters for an electron accelerator, or for even simpler tasks that can be run and optimized just in time with a standard laptop at hand. \\

Our main contributions is 1.) comparing how the log-likelihood affects the angle-difference in the real projection matrix, and the found matrix matrix, 2.) an extensive analysis of current popular methods including strengths and shortcomings, 3.) a short analysis on how dimensionality reduction techniques can be used for feature selection, and 4.) a novel algorithm called "BORING", which allows for a simple fallback mechanism if the matrix identification fails, as well as taking into consideration "passive" subspaces which provide small perturbations of the function at hand. \\

The main features of BORING are 1.) the possibility to identify the subspace (unlike most other optimization algorithms), and 2.) to provide a much lower penalty to identify the subspace if identification fails, as optimization is still the primary goal.

\end{abstract}


\tableofcontents


\printnomenclature

\mainmatter

\chapter{Introduction}

Tuning hyperparameters is considered a computationally intensive and tedious task, be it for neural networks, or complex physical instruments such as free electron lasers.
Users for such applications could benefit from a 'one-click-search' feature, which would find optimal parameters in as few function evaluations as possible.
This project aims to find such an algorithm which is both efficient and holds certain convergence guarantees.
We focus our efforts on Bayesian Optimization (BO) and revise techniques for high-dimensional BO. \\

In Bayesian Optimization, we want to use a Gaussian Process to find an optimal parameter setting $\mathbf{x^*}$ that maximizes a given utility function $f$.
The procedure consists of 1. generating such a surrogate surface, using Gaussian Processes, and 2. applying an acquisition function.
The acquisition function allows for a tradeoff between exploitation of the best found global maximum so far, and exploration - searching for a configuration $x$ whose neighborhood was not picked yet.\\

\subsection{Main Contributions}

Our main contributions are:

\begin{enumerate}
\item We conduct a systematic analysis of the state of the art and identify shortcomings and strengths of some popular methods   .
\item We define a new algorithm called "BORING," which is supposed to take into consideration small perturbations that could yield marginal improvement over standard algorithms.
BORING also provides a more robust framework than some other methods concerning the accuracy of identified subspaces.
One can say that BORING provides a fallback mechanism when the matrix identification fails, and otherwise only adds a small penalty concerning regret.
\item We demonstrate that the log-likelihood loss is not a single best metric to optimize over, and demonstrate its correlation to the angle-difference between embeddings for three different types of functions.
One cannot rely upon that the decreasing log-likelihood will decrease the angle-difference between the real embedding matrix and the found embedding matrix. However, the log-likelihood provides a good heuristic for this task for most functions.
\item We conduct a short analysis to what extent linear dimensionality reduction techniques can be used for feature selection.
\end{enumerate}

We will talk about the problems and possible solutions for the task at hand in the next section.


\chapter{Background}  

\ifpdf
    \graphicspath{{Chapter1/Figs/Raster/}{Chapter1/Figs/PDF/}{Chapter1/Figs/}}
\else
    \graphicspath{{Chapter1/Figs/Vector/}{Chapter1/Figs/}}
\fi

\section{Bayesian Optimization in high dimensions} 

Many technical problems can be boiled down to some flavor of black box optimization. 
Such problems include neural architecture search \citep{BayesianOptimizationNAS}, hyper-parameter search for neural networks, parameter optimization for electron accelerators, or drone parameter optimization using safety constraints \citep{berkenkamp17saferl}. \\

Bayesian optimization methods are a class of sequential black-box optimization methods.
A surrogate function surface is learned using a Gaussian prior, and a Gaussian likelihood function.
Combining the prior and the likelihood results in the Gaussian posterior, which can then be used as a surface over which optimization can take place, with the help of a chosen acquisition function. \\

Bayesian optimization is a method that has increasingly gained attention in the last decades, as it requires relatively few points to find an appropriate response surface for the function over which we want to optimize over.
It is a sequential model based optimization function, which means that we choose the best point $x^*_i$ given all previous points $x^*_{i-1}, x^*_{i-2}, \ldots, x^*_{0}$.
Given certain acquisition functions, it offers a good mixture of exploration and exploitation from an empirical standpoint \citep{BOIncreasingPopularityEmpirically}. \\

However, as machine learning algorithms and other problems become more complex, Bayesian optimization needs to cope with the increasing number of dimensions that define the search space of possible configurations.
Because BO methods lose effectiveness in higher dimensions due to the curse of dimensionality, this work explores Bayesian optimization methods that improve the optimization performance in higher dimensions.
Finally, we propose a novel method that can take into consideration small perturbations of the function we want to optimize over.

\section{Gaussian Processes}
Bayesian Optimization (BO) aims at using a Gaussian Process as an intermediate representation to find an optimal parameter setting $\mathbf{x^*}$ that maximizes a given utility function $f$.\\

Assume we have observations $ \mathcal{Y} = \{ y^{(1)}, \ldots, y^{(N)} \}$, each evaluated at a point $ \mathcal{X} = \{  \mathbf{x}^{(1)}, \ldots, \mathbf{x}^{(N)} \}$.
The relationship between the observations $y$ and individual parameter settings $\mathbf{x}$ is $y = f \left( \mathbf{x} \right) + \epsilon$ where $\epsilon \sim  \mathcal{N} \left( 0, \sigma^2_n \right)$. Any quantity to be predicted has a subscript-star (e.g. $y_*$ is the function evaluation we want to predict).\\

In it's simplest form, a Gaussian Process is described by the following equation:

\begin{equation}
\begin{pmatrix} y \\
y_* \end{pmatrix} \sim N\Biggl(\mu,\begin{pmatrix} K & K^T_*\\
 K_* & K_{**} \end{pmatrix}\Biggr),
\end{equation}

Where $\mu$ is a mean function, $K = \text{kernel}(\mathbf{X}, \mathbf{X})$, $K_* = \text{kernel}(\mathbf{x_*}, \mathbf{X})$ and $K_{**} = \text{kernel}(\mathbf{x_*}, \mathbf{x_*})$.
Any new point $y_*$ is predicted given all previously sampled points $y$ by estimating the probability $ p(y_*|y) \sim N(K_*K^{-1}y,K_{**}-K_*K^{-1}K'_*) $\\

An acquisition function can then use these results, that described where to best sample the points next.
Some popular acquisition functions include GP-UCB, Most probable improvement (MPI) and Expected Improvement (EI).
The choice of the acquisition function has a significant influence on the performance of the optimization procedure.\\

In the following, I provide a short derivation of the core formulae used for Gaussian Processes.

\subsection{Derivation of the Gaussian Process Formula}
The \textbf{prior} for the Gaussian Process is the following (assuming the commonly chosen 0-mean-prior).

\begin{equation}
u \sim GP(0, k(x, x'))
\end{equation}

$u$ is a random variable following a Gaussian Process distribution and its probability distribution is given by a normal distribution:

\begin{equation}
p(u) = N ( 0, K )
\end{equation}

Additional data $ X = \{ (x_1, y_1), \ldots, (x_n, y_n) \} $ can be observed.
The Gaussian Process incorporates an error term that takes into consideration possible noise in the measurement of the experiment.
Thus, $y$ has some noise term $\epsilon$ such that $y = u(x) + \epsilon$.
The common assumption is taken that $\epsilon$ is normally distributed around $0$ with $\sigma_s$ standard deviation.
Given the sampled datapoints, and the inherent noise that these datapoints have, the \textbf{likelihood} of the Gaussian Process can be represented as follows:

\begin{equation}
p(y | x, u) = N (u, \sigma_s^2 I)
\end{equation}

Given the prior and likelihood of the Gaussian Process, the \textbf{posterior} of the Gaussian Process can be derived by simple application of Bayes rule.

\begin{align}
p(u | x, y) &= \frac{ p(y | x, u) p(u) }{p(y | x)}
& = N( K(K +\sigma^2 I)^{-1}y, \sigma^2 (K + \sigma^2 I)^{-1} K )
\end{align}

From the above posterior, we now want to predict for an arbitrary $x_*$ the function value $y_*$.
Predicting $y_*$ for every possible $x_*$ in the domain results in the surrogate response surface that the GP models. \\

We assume that the value $y_*$ we want to predict is also distributed as a Gaussian probability distribution. 
Because the $y_*$ that we want to predict relies on all the values collected in the past (which are again normally distributed), the probability distribution can be modelled as jointly Gaussian:

\begin{equation}
\begin{pmatrix} y \\
y_* \end{pmatrix} \sim N\Biggl(\mu,\begin{pmatrix} K & K^T_*\\
 K_* & K_{**} \end{pmatrix}\Biggr),
\end{equation}

To compute this equation, we use the results from Murphy's textbook \citep{Murphy} pages 110 to 111 to make inference in a joint Gaussian model.
The transition from the Gaussian Process, to Bayesian Optimization lies in finding the argmax of the above term.
This argmax can be sequentially estimated using the following estimation functions.

\section{Acquisition Functions}

Given the above formula for the posterior mean $\mu$ and the poster variance $\sigma^2$, Bayesian Optimization makes use of an acquisition function.
The following is a summary of the most popular acquisition functions in recent literature.
A good summary is given by \citep{AcquisitionFunctionsMaximizing}.

\subsection{Upper Confident Bound (UCB)}
\citep{UCBRegretProof} shows a proof, that for a certain tuning of the parameter $\beta$, the acquisition function has asymptotic regret bounds.

The upper confidence bound allows the user to control exploitation and exploration through a parameter $\beta > 0$, which can be chosen as specified in \citep{UCBRegretProof} to offer regret bounds.
In addition to that, GP-UCB shows state-of-the-art empirical performance in numerous use-cases\citep{Djolonga2013}.

\begin{equation}
UCB(x) = \mu(x) + \sqrt{ \beta } \sigma(x)
\end{equation}

Here, the functions $\mu$ and $\sigma$ are the predicted mean and variance of the Gaussian Process Posterior.

\subsection{Probability of Improvement (PI)}
The (maximum) probability of improvement \citep{AcquisitionFunctions} always selects the points where the mean plus uncertainty is above the maximum explored function threshold. 
The downside to this policy is that this leads to heavy exploitation.
However, the intensity of exploitation can be controlled by a parameter $\xi > 0$.

\begin{align}
    PI(x) & = P( f(x) \geq f(x^+) + \xi ) \\
    & = \Phi ( \frac{\mu(x) - f(x^+) - \xi}{\sigma(x)}  ) 
\end{align}

\subsection{Expected Improvement (EI)}
As an improvement to the maximum probability of improvement, the expected improvement takes into consideration not only the probability that a point can improve the maximum found so far.
The EI also takes into account the magnitude by which it can improve the maximum function value \citep{AcquisitionFunctions}.
As in MPI, one can control the rate of exploitation by setting the parameter $\xi > 0$, which was introduced by \citep{Lizotte2008}.

\begin{align}
    EI(x) =
    \begin{dcases}
        ( \mu (x) - f(x^+) - \xi) \Phi(Z) + \sigma (x) \phi (Z) & \text{ if } \sigma (x) > 0 \\
        0 & \text{ if } \sigma (x) = 0
    \end{dcases} \\
\end{align}

where

\begin{equation}
    Z = \frac{\mu (x) - f(x^+) - \xi}{\sigma(x)}
\end{equation}

and where $\phi$ denotes the PDF, and $\Phi$ denotes the CDF of the standard normal distribution respectively. \\

Given one of the above acquisition functions, one can then use an optimizer such as $LBFGS$ or monte carlo sampling methods, to find an approximate global maximum of the respective function.
The combination of Gaussian Processes and Acquisition function together result in a Bayesian Optimization algorithm, which has a prior assumption about the function to be learned and uses data-samples to create a likelihood to refine the posterior of the initial function assumption further.

\section{Resources}
For the experiments and code basis, most of our functions rely on the Sheffield Machine Learning - GPy library \citep{gpy2014}.
In addition to that, I use the febo framework developed by Johannes Kirschner from the Learning and Adaptive Systems group at ETH Zurich.


\chapter{Related Work}

\ifpdf
    \graphicspath{{04_Chapter2/Figs/Raster/}{04_Chapter2/Figs/PDF/}{04_Chapter2/Figs/}}
\else
    \graphicspath{{04_Chapter2/Figs/Vector/}{04_Chapter2/Figs/}}
\fi

This section covers current methods solving Bayesian Optimization in high dimensions that are considered state-of-the-art.
This document is not an exhaustive review. 
For each algorithm, I discuss the effectiveness with regards to the dataset or function that the respective paper evaluates on.

Unless specified differently, I use the following terminology during my discussion.

\begin{enumerate}
\item $f$ is the real function to be optimized. Often I will use the term approximate, as many algorithms rely on a surrogate approximation of the function $f$ such that the global optimum can be found.
\item Assuming the function $f$ is of the form $f(x) = y$ with $x \in \mathbf{R}^D$, where $D$ denotes the dimensionality of $x$, then we define the optimized value of $f$ to be $x^* = \arg \max_{x} f(x)$.
\item $g$ and any subscripted or superscripted derivative of $g$ is part of the surrogate function of $f$.
\item Anything that has a "hat" on (caret symbol on f is $\hat{f}$ ), refers to an empirical estimate. 
$\hat{f}$ would be an empirical estimate given datapoints $\mathit{D}$ of $f$.
\end{enumerate}

I will focus on three categories of 33Bayesian Optimization algorithms: Algorithms that make use of a projection matrix, algorithms that exploit additive substructures and "additional approaches" that are uncategorized.

\section{Projection matrix-based algorithms}

In my work, I focus on algorithms that optimize the black box function $f$ by using a lower-dimensional projection.
I proceed with discussing some algorithms that I have implemented for my experiments in the next subsection (all except the algorithm "Active learning of linear subspaces").
For this family of algorithms, the approximation is a function of $f(x) \sim g(x; A)$, where the properties of $A$ are algorithm-specific.
The following descriptions aim at giving a brief overview of the methods at hand. 
I refer the curious reader to the individual paper for a more detailed and formal description of the respective topic.

\subsection{Active learning of linear subspaces}

Although the active learning of linear subspaces is not a BO algorithm, it is still relevant, which is why I am presenting it here.

\begin{algorithm}
\caption{Simultaneous active learning of functions and their linear embeddings (pseudocode) :: Active learning of linear subspace \citep{Garnett2013}}

\begin{algorithmic} 
\REQUIRE $d, D;$ kernel $\kappa$, mean function $\mu$; prior $p(R)$ 
\STATE $X \leftarrow \emptyset$
\STATE $Y \leftarrow \emptyset$

\WHILE{budget not depleted}
\STATE $ q(R) \leftarrow \text{LAPLACEAPPROX}( p(R | X, Y, \kappa, \mu) ) $
\STATE $ q(f) \leftarrow  APPROXMARGINAL( p(f | R), q(R)) $
\STATE $ x_* \leftarrow OPTIMIZEUTILITY( q(f), q(R) )$
\STATE $ y \leftarrow OBSERVE( f( x_* ) ) $
\STATE $ X \leftarrow [X; x_*] $
\STATE $ Y \leftarrow[Y; y_*] $
\ENDWHILE

\RETURN $q(R), q(f)$
\end{algorithmic}

\end{algorithm}

\citep{Garnett2013} The assumption of this algorithm is that $f$ depends only on $ x := uR^T $ with $ R \in \mathbf{R}^{d \times D}$, $ u \in \mathbf{R}^d $ and where $d << D$. 
The algorithm learns a projection matrix $R$ and the surrogate function $g(u)$, with $f(x) \sim g(u) $. \\

The \textbf{Laplace approximation} for $R$ is using the mode of the probability distribution $\log P (R | D) $ as a mean, and the covariance is taken as the inverse Hessian of the negative logarithm of the posterior evaluated at the mean of the distribution.
Together, this describes the probability distribution $p(R|X, Y, \kappa, \mu )$, where $\mu$ is the mean function, and $\kappa$ is the covariance function.\\

The \textbf{approximate marginal} subroutine is a novel method proposed in the paper that integrates over the different parameters in the paper.
This marginal approximation does a local expansion of $q(x_* | \theta) $ to $p(x_* | D, \theta)$. \\

The \textbf{sequential optimization of utility} (choice of next best point) is done using Bayesian Active Learning by disagreement, where the utility function is the expected reduction in the mutual information, as opposed to uncertainty sampling reducing entropy, which is not well defined for all values. \\

The metrics used in this paper are negative log-likelihoods for the test points and the mean symmetric kullback leiber divergence between approximate and true posteriors.
The proposed method always outperforms the naive MAP method for the presented functions close to a factor of 2 for both loss functions.
Tests are conducted on a real, and synthetic dataset with up to $D = 318$ and selecting $N = 100$ observations. 

\subsection{Random embeddings (REMBO)}
REMBO is an algorithm that allows the optimizer to search in a smaller search space.
The result of the optimized value $x^*$ is then projected to the higher dimensional space, to retrieve the actual optimized argmax value of the function $f$. 
More specifically, REMBO proposes the following model for Bayesian Optimization: \\

\citep{Wang2013} Let $x \in \mathbb{R}^D$ and $y \in \mathbb{R}^d$. Assume, that $f(x) = g(Ay)$. We can generate $A \in \mathbb{R}^{D \times d}$ by randomly generating this matrix.
The space over which the user searches, as such, is $d$-dimensional.
This implies that REMBO is more efficient in sampling data points, and learning the structure of a neighborhood in the high-dimensional space (by using the neighborhood of the low-dimensional space). \\

\begin{figure}[H]
    \centering
        \includegraphics[width=\textwidth]{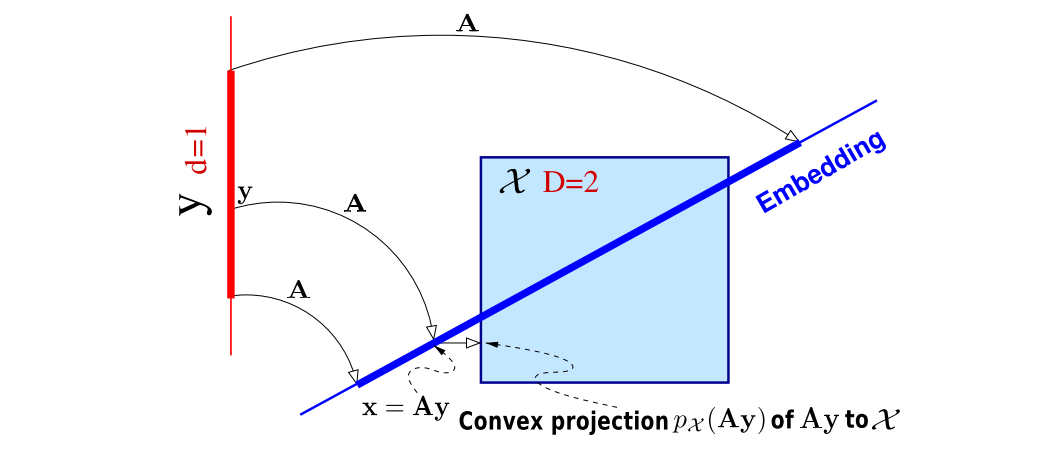}
        \caption{Parabola Original}
        \label{fig:gull}
    \caption{
    Source \citep{Wang2013}: Embedding from $d = 1$ into $D=2$.
    The box illustrates the 2D constrained space $\mathbf{X}$, while the thicker red line demonstrates the 1D constrained space $\mathbf{Y}$.
    Note that if $A \times y$ is outside of $\mathbf{X}$, it is projected onto $\mathbf{X}$ using a convex projection.
    The set $\mathbf{Y}$ must be chosen large enough so that the projection of its image, $A \times y $ with $y \in \mathbf{Y}$, onto the effective subspace (vertical axis in this diagram) covers the vertical side of the box.
    }\label{fig:animals}
\end{figure}

The elegance of REMBO lies in the fact that the matrix $A$ can be chosen as a random matrix.
Through an experiment, the authors argue, that the chance of getting a bad projection matrix has a certain threshold if the optimization domain's parameters are well-chosen. \\

I now proceed with a more formal treatment of the intuitive concept explained above: \\

\cite{Wang2013}
A function $f : \mathbf{R}^D \rightarrow \mathbf{R}$ is said to have effective dimensionality $d_e$ (where $d_e < D$), if there exists a linear subspace $\mathcal{T}$ of dimension $d_e$ such that for all $ x_\top \in \mathcal{T} \subset \mathbf{R}^D $ and $x_\perp \in \mathcal{T_\perp} \subset \mathbf{R}^D $, we have $ f(x) = f(x_\top +x_\perp ) = f(x_\top)$.
$\mathcal{T^\perp}$ is the orthogonal complement of $\mathcal{T}$.

Assume $ f : \mathbf{R}^D \rightarrow \mathbf{R} $ has effective dimensionality $d_e$.
Given a random matrix $ \mathbf{A} \in \mathbf{R}^{D \times d} $ (where $d \geq d_e$) with independent entries sampled from $ \mathcal{N}(0, 1) $.
For any $ x \in \mathbf{R}^D $, there exists a $y \in \mathbf{R}^d $ such that $ f(x) = f(\mathbf{A} y ) $.
The user now only need to optimize over all possible $y \in \mathbf{R}^d$, instead of all possible $x \in \mathbf{R}^D $. \\

\begin{figure}[H]
    \centering
        \includegraphics[width=\textwidth]{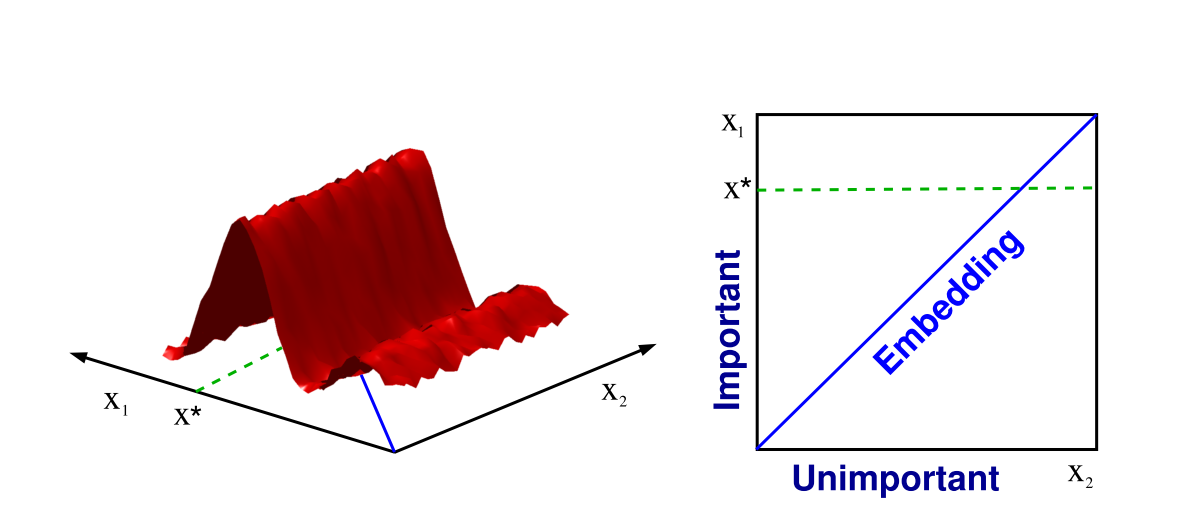}
        \caption{Parabola Original}
        \label{fig:gull}
    \caption{
    Source \citep{Wang2013}: This function in D=2 dimesions only has d=1 effective dimension.
    Hence, the 1-dimensional embedding includes the 2-dimensional function’s optimized value $x^*$. 
    It is more efficient to search for the optimum along the 1-dimensional random embedding than in the original 2-dimensional space.
    }\label{fig:animals}
\end{figure}

REMBO has a reasonably high probability of failing (more than 20\% in the experiments conducted in the paper) by choosing an $A$ that lies roughly orthogonal the directions of highest change in the function $f$.
The authors propose interleaved runs, where for each $k$'the point selection, a differently sampled orthogonal random matrix is chosen. 
The probability of generating a bad embedding lower with higher $k$.
However, each $k$ has a set of observations $X$ which is not shared across the different interleaved runs. \\
Extensions to REMBO include \citep{RemboExtension}.

\subsection{GPs with builtin dimensionality reduction}
I put most emphasis on this algorithm during my experiments. 
As such, I will be more detailed with the description of this algorithm.
During the rest of this dissertation, I will refer to this algorithm as "Tripathy's method", or "Tripathy's algorithm".\\

\citep{Tripathy} This algorithm assumes that  $f(x) \sim g( \mathbf{W}^T y)$ where $ \mathbf{W} \in \mathbb{R}^{D \times d} $ and $D >> d$.
$ \mathbf{W} $ again has the property that $\mathbf{W^T}  \times \mathbf{W} = I$ where $I \in \mathbf{R}^{d \times d}$ denotes the identity matrix in $d$ dimensions. 
This algorithm does not require gradient-information.
This makes it easier to implement, and more robust to noise according to the authors of this paper. \\

I refer to the set of kernel parameters and the GP noise variance as GP-hyperparameters.
I refer to the projection matrix $ \mathbf{W} $ as the projection matrix.

The GP noise variance, kernel parameters, and  $ \mathbf{W} $ can be found iteratively.
The main idea of the algorithm is to first fix both the kernel parameters and the GP noise variance, and identify $ \mathbf{W} $.
Then, we fix $ \mathbf{W} $ and train over the GP-hyperparameters.
This procedure is repeated until the change of the log-likelihood between iterations is below some $ \epsilon_l $, or if the maximum number of steps of optimization is reached.
We repeat this procedure many times (as dictated by the number of restarts).
Empirically, the probability of finding a good projection increases with the number of restarts, and the maximum number of iterations the algorithm is allowed to take during the optimization of the joint parameter set $\mathbf{W}$ and the hyperparameters (as given in the next paragraphs).\\

\paragraph{Formal description of the algorithm.}
The quantities of interest are the mean of the function $\mu_f$ and the variance $\sigma^2_f$.
The authors assume that w.l.o.g., the search space of the projection matrices can be restricted to matrices on the Stiefel manifold.
The argument for this is that the projected subspace does not change, merely the representation of the subspace.
Here, the family of orthogonal matrices of dimension $d \times D$ is denoted by $\mathbf{W} \in V_d(\mathbb{R}^D) $.
This quantity is also known as the \textbf{Stiefel manifold} \citep{StiefelBayesianInference} \citep{StatisticsStiefelIntro} \citep{StiefelNonparametric}, where $d$ is the found effective dimension of the function, and $D$ is the real input dimension to the function. \\

\subsubsection{The Matern32 Kernel}
The optimization processes in the paper use the Matern-32 kernel function.
This function has two input vectors $a$ and $b$.

\begin{align}
K(a,  b, \theta) = s^2 \left( 1 + \sqrt{3} \sum_{i=1}^l \frac{(a_i - b_i)^2}{ \textit{l}_i} \right) exp\left( - \sqrt{3} \sum_{i=1}^l \frac{(a_i - b_i)^2}{ \textit{l}_i} \right)
\end{align}

Because I want to avoid numerical testing and implementation, I use the derivative as provided in the GPy library.
The $s, l_1, \ldots, l_l $ are hyper-parameters of the kernel, referred to as the kernel-variance, and the kernel-lengthscales. 
The concatenated vector of all these kernel hyperparameters is denoted by $\phi$. \\

The only modification made to this kernel is the additional parameter $W$:

\begin{equation}
k_{AS} : \mathbb{R}^D \times \mathbb{R}^D \times V_d(\mathbb{R}^D) \times \phi \rightarrow \mathbb{R} \\
\end{equation}
\text{where the kernel has the form}
\begin{equation}
k_{AS} (x, x'; W, \phi) = k_d(W^T x, W^T x'; \phi)
\end{equation}

\subsection{Overview of the algorithm}

In the following, I will explain the individual steps of the algorithm.
The shown two steps are repeated until the change of the log-likelihood reaches a certain threshold.
This algorithm is repeated a number of restarts, as the initial sample of $W$ plays a significant role in finding a well converged embedding.

\subsubsection{Step 1.: Determine the active projection matrix W}
In this step, the algorithm optimizes $W \in V_d(\mathbb{R}^D)$ while keeping the kernel hyperparameters $\phi$ and the GP noise variance $s_n$ fixed.

To simplify calculations later, we define a function $F$, where all parameters but $W$ are fixed. 
The other parameters are determined from previous runs, or are randomly initialized:

\begin{align} \label{Eq:LogLikelihoodF}
F(W) &:= \mathcal{L}(W, \phi, s_n; X, y) \\
& = \log p(y | X, W, \phi, s_n) \\
& =  -\frac{1}{2} (y - m)^T (K + s_n^2 I_N)^{-1} (y - m) -\frac{1}{2} \log|K + s_n^2 I_N| -\frac{N}{2} \log 2 \pi   \\
\end{align}

where $\phi, s_n; X, y$ are fixed, and $m$ is the prior mean function, which is 0 in our specific case.

To optimize over the loss function, the algorithm defines the derivative of $F$ with regards to each individual element of the weights-matrix:

\begin{align}
\nabla_{w_{i,j}} F(W) &:= \nabla_{w_{i,j}} \mathcal{L}(W, s_n; X, y) \\
& = \frac{1}{2} \text{tr} \left[ \{ (K + s_n^2 I_N)^{-1} (y-m) \left( (K + s_n^2 I_N)^{-1} (y-m) \right)^T - (K + s_n^2 I_N)^{-1} \} \nabla_{w_{i,j}} (K + s_n^2 I_N) \right]
\end{align}

both these functions depend on the kernel $K$, and it's derivative $\nabla_{w_{i,j}} K$. \\

A more sophisticated algorithm is used to optimize over $F$, that resembles iterative hill-climbing algorithms.
First, the paper defines the function whose output is a matrix in the Stiefel manifold.

\begin{equation}\label{Eq:TauFunction}
\gamma(\tau; W) = (I_D - \frac{\tau}{2} A(W) )^{-1} (I_D + \frac{\tau}{2} A(W) ) W
\end{equation}

where $W$ is a fixed parameter, and $\tau$ is the variable which modifies the direction that we move on in the Stiefel manifold and with

\begin{equation}
A(W) = \nabla_{W} F(W) W - W ( \nabla_{W} F(W) )^T
\end{equation}

One iteratively chooses fixed $W$ and does a grid-search over $\tau \geq 0$ such that at each step, the log-likelihood $\mathcal{L}$ is increased.
The grid-size is between 5 and 100 for our experiments.

\subsubsection{Step 2.: Optimizing over GP noise variance and the kernel hyperparameters}

We determine the hyperparameters by optimizing over the following loss function, where $X$ are the input values, $Y$ are the corresponding output samples. $\phi$ is the vector of the kernel hyperparameters and $s_n$ the GP noise variance. \\

One keeps the $W$ fixed (either by taking $W$ from the last iteration, or freshly sampling it), and then defines the loss function.

\begin{equation}
    L(\phi) = \mathcal{L}(W, \phi, s_n; X, y) 
\end{equation}

To optimize this loss function, a simple optimization algorithm such as $L-BFGS$ is used to individually maximize each element of the hyperparameter vector with regards to the log-likelihood.
This is done for a maximum number of steps, or until the change of improvement becomes marginal. \\

\subsubsection{Additional details}
Because initialization is a major factor in this algorithm, these steps are iteratively applied for many hundred steps.
There are also many tens or hundreds of restarts to ensure that the search on the Stiefel manifold results in a suitable local optimum, and does not get stuck on a flat region with no improvement.
This algorithm is very sensitive to the initially sampled parameter $W$.

\subsubsection{Identification of active subspace dimension }
One does not know the real active dimension of the optimization problem at hand.
As such, the proposed to apply the above algorithms iteratively by increasing the selected active dimension $d$.
The moment where the relative change between the best-found matrix between two iterations is below a relative threshold $\epsilon_s$, the previous active dimension is chosen as the real active dimension. 
The algorithm identifies the loss for each possible active dimension.
It then chooses a dimension, where the relative difference to the previous loss (of the one-lower dimension) is below a certain threshold.

\section{Algorithms that exploit additive substructures}

Functions with additive substructures can be decomposed into a summation over subfunctions, such that
$ f(x) \sim g_0(x) + g_1(x) + \ldots g_n(x) $ where each $g_i$ may operate only on a subset of dimensions of $x$.

\subsection{Independent additive structures within the target function}

\citep{Gardner2017} Assume that $f(x) = \sum_{i=1}^{ |P| } f_i (x[P_i] )$, i.e. $f$ is fully additive, and can be represented as a sum of smaller-dimensional functions $f_i$, each of which accepts a subset of the input-variables.
The kernel also results in an additive structure: $f(x) = \sum_{i=1}^{ |P| } k_i (x[P_i], x[P_i])$.
The posterior is calculated using the Metropolis-Hastings algorithm.
The two actions for the sampling algorithm are 'Merge two subsets', and 'Split one set into two subsets'.
$k$ models are sampled, and we respectively approximate $p(f_* | D, x^*) = \frac{1}{k} \sum_{j=1}^{k} p( f(x^* | D, x, M_j) )$, where $M_j$ denotes the partition amongst all input-variables of the original function $f$.

\section{Additional approaches}

\subsection{Elastic Gaussian Processes}

\citep{Rana2017} Use a process where the space is iteratively explored.
The key insight here is that with low length-scales, the acquisition function is flat, but with higher length-scales, the acquisition function starts to have significant gradients.
The two key-steps is to 1.) additively increase the length-scale for the Gaussian process if the length-scale is not maximal and if $|| x_{init} - x^* || = 0$.
And 2.) exponentially decrease the length-scale for the Gaussian process if the length-scale is below the optimum length-scale and if $|| x_{init} - x^* || = 0$.

\subsection{High dimensional Gaussian bandits}

\citep{Djolonga2013} This model assumes that there exists a function $g : \mathbf{R}^k \implies [0, 1]$ and a matrix $A \in \mathbf{R}^{d \times D}$ with orthogonal rows, such that $f(x) \sim g(Ax) $. Assume $g \in \mathcal{C}^2$. 

The algorithm identifies $A$ by the 

\begin{algorithm}
\caption{The SI-BO algorithm \citep{Djolonga2013}}

\begin{algorithmic} 
\REQUIRE $m_X, m_{\Phi}, \lambda, \epsilon, k$, oracle for the function $f$, kernel $\kappa$ 

\STATE $C \leftarrow m_X $ samples uniformly from $\mathbb{S}^{d-1}$

\FOR{$ i \leftarrow 1$ to $m_X$}
\STATE $\Phi_i \leftarrow m_{\Phi}$ samples uniformly from $\{ -\frac{1}{\sqrt{m}}, \frac{1}{\sqrt{m}} \}^k$
\ENDFOR

\STATE $ y \leftarrow $ Compute $y$ using Equation 1 from \citep{Djolonga2013}

\STATE select $z_i$ according to a UCB acquisition function, evaluate $f$ on it, and add it to the data samples found so far

\end{algorithmic}

\end{algorithm}

The SI-BO algorithm consists of a subspace identification step, and an optimization step using GP-UCB.

The \textbf{subspace identification} is treated as a low-rank matrix recovery problem as presented in \citep{CevherSubspaceIdentificationKrause}.

\subsection{Bayesian Optimization using Dropout}

\citep{Li2018} propose that the assumption of an active subspace is restrictive and often not fulfilled in real-world applications.
They propose three algorithms, to iteratively optimize amongst particular dimensions that are not within the $d$ 'most influential' dimensions: 1.) Dropout Random, which picks dimensions to be optimized at random, 2.) Dropout copy, which continuous optimizing the function values from the found local optimum configuration, and 3.) which does method 1. with probability $p$, and else method 2.
The $d$ 'most influential' dimensions are picked at random at each iteration. \\

Additional works I explored as background research or additional methods include \citep{KernelGibbsSampler}, \citep{VirtualVsReal}, \citep{SensorPlacement}, \citep{BatchedBO}, \citep{GPforML}.

\chapter{Analysis of the current state of the art}

\ifpdf
    \graphicspath{{Chapter3/Figs/Raster/}{Chapter3/Figs/PDF/}{Chapter3/Figs/}}
\else
    \graphicspath{{Chapter3/Figs/Vector/}{Chapter3/Figs/}}
\fi

\section{Shortcomings of current methods}
I will enumerate select models from the section "related work," and will shortly discuss what the shortcomings of these models are:

\paragraph{REMBO} is a BO algorithm which finds $y^*$ in lower dimensions This $y^*$ is then projected to the higher dimension, which has a chance of being the optimal point $x^* = \arg \max_{x} f(x)$. 

\begin{itemize}

\item \textbf{Suitable choice of the optimization domain:} REMBO is not robust, as there is a considerable probability that no suitable subspace will be found. 
Empirically, the choice of the optimization domain profoundly affects the duration and effectiveness of the optimization.
I have found that the proposed optimization domain $ \left[ -\sqrt{d}, \sqrt{d}  \right]^d $ is not well chosen for smaller environments, such as the Camelback function embedded in 5 dimensions.
In any case, this is a very sensitive hyperparameter.

\item \textbf{Identification of subspace:} In some settings, including optimization with safety constraints, knowing the subspace that the model projects to is advantageous. 
REMBO is an implicit optimizer, in that it does not find any subspace, but optimizes through a randomly sampled matrix.

\item \textbf{Probability of failure:} REMBO has a relatively high probability of failure. 
The authors propose that restarting REMBO multiple times would allow for a proper optimization domain to be found, which leads to interleaved runs. 

\end{itemize}

\paragraph{Active subgradients} can be a viable option if we have access to the gradients of the problem, from which we can learn the active subspace projection matrix in the manner by using that gradient matrices.

\begin{itemize}

\item \textbf{Access to gradients:} For optimization algorithms, the function we want to optimize over is usually a black-box function.
Practically, many black-box functions do not offer access to gradient information.
To approximate the gradients using sampled points, this would require a high number of data points per dimension.
In addition to that, these points would have to be evenly distributed, such that the gradients can be adequately estimated for more than one region.

\item \textbf{Robustness to noise:} According to \citep{Tripathy}, methods that approximate gradients and use this gradient information to approximate a subspace are very sensitive to noise.
Depending on the application, this can make the algorithm ineffective as it is not robust to small variations in the response surface.

\end{itemize}

Given the nature of real-world data, approximating the active subspace using the gradients of the data-samples is thus not a robust, and viable option.

\paragraph{Bilinois et al.} (which I also refer to as "Tripathy's method" and refers to the "GP with builtin dimensionality reduction) argue that their method is more robust to real-world noise. 
It also does not rely on gradient information of the response surface.
Bilinois's method allows for a noise-robust way to identify the active subspace.

\begin{itemize}

\item \textbf{Duration of optimization:} In practice, Bilinois's method takes a long time, especially if the dimensions or the number of data-points are high. This is due to the high number of matrix multiplications. 
Especially for easier problems, it is often desirable if the running time of optimizing for the next point is a few minutes or seconds, rather than hours.

\item \textbf{Efficiency:} In practice, Tripathy's method relies on a high number of restarts.
From our observations, the number of steps to optimize the orthogonal matrix becomes relevant as the number of dimensions grow.
Given the nature of accepting any starting point, it does not allow for a very efficient way to search for the best possible projection matrix.
A more efficient way to search all possible matrices - by incorporating heuristics for example - would be desirable.

\item \textbf{Insensitive to small perturbations:} Although Tripathy's model finds an active subspace, it completely neglects other dimensions which could allow for small perturbations to allow for an additional increase the global optimum value.
Although we can control to what extent small perturbations should be part of the active subdimension, one usually wants to choose a significant cutoff dimension, but still, incorporate additional small perturbations without sacrificing the effectiveness of the projection.

\end{itemize}

\section{Evaluation methods}
In the following sections, we will discuss and show how we can improve on the shortcomings of the above methods.
Because practicality is vital in our method, we will use synthetic functions to measure the efficiency of our method.

Some terms that allow us to measure the performance of a Bayesian optimization algorithm or a GP surrogate function include:

\begin{itemize}
\item Test if the expectation $$ E[ f(A x) - \hat{f}(\hat{A} x) ] $$ decreases / approaches zero (for methods that identify a projection matrix).
Often, the root mean square error is a good empirical approximate of this quantity:
\begin{equation}
RMSE = \sqrt{ \frac{1}{T} (\sum_{t=1}^{T} f(A x_t) - \hat{f}(\hat{A} x_t))^2 }
\end{equation}

The log-likelihood estimate is also an estimate which tests this value for the training data.

\item For optimization problems, one is often interested in the quantity of cumulative regret.
Regret is defined as the difference between the best found function value so far, minus the function value chosen at this timestep $t$ \citep{RegretDef}.

\begin{equation}
R_T = \frac{1}{T} \sum_{t=1}^{T} \max_x f(x) - f(x_t)
\end{equation}

The cumulative regret sums all the entire episode of the run.
This is a measure of how fast an optimizer can learn the optima of a function.

\item Check if the test log-likelihood w.r.t. to the GP decreases for functions that are provided by a finite number of data-points.

\item Check if the angle between the real projection matrix and the found projection matrix decreases, as given in \citep{AngleMeasurement}. 

\begin{align}
dist(A, B) &= \left\Vert A A^T - B B^T\right\Vert_2 \\
& = sin( \phi )
\end{align}

where $ A, B \in \mathbf{R}^{D \times d}  $
\end{itemize}

\subsection{Synthetic Datasets} \label{syntheticFunction}
\paragraph{5 dimensional function with 2 dimensional linear embedding}

One can evaluate synthetic functions at any point.
This allows analyzing the regret of a specific BO algorithm.
The following synthetic functions cover different use cases.

\paragraph{2D to 1D}: A simple Parabola which is embedded in a 2D space.
This function is meant as a sanity check.
In other, this function is simple, and any algorithm taken into consideration should be able to find an effective subspace for this simple function.
\paragraph{3D to 2D}: The Camelback function which is embedded in a 3D space.
This checks how tight the model can approximate the 2D subspace, or if a 3D UCB performs better than our models.
\paragraph{5D to 2D}: The Camelback function which is embedded in a 5D space.
This checks if more complicated models can be found within higher dimensional spaces.
\paragraph{5D to 2D}: The Sinusoidal Exponential function which is embedded in a 5D space.
This is a function which consists of two additive subfunction. 
The first one has high perturbations amongst a given axis. 
The second has low perturbations amongst an axis orthogonal to the first one.
This function allows for benchmarking to what extent the algorithm accounts for small perturbations.

\chapter{A New Model}

\ifpdf
    \graphicspath{{Chapter4/Figs/Raster/}{Chapter4/Figs/PDF/}{Chapter4/Figs/}}
\else
    \graphicspath{{Chapter4/Figs/Vector/}{Chapter4/Figs/}}
\fi

We now propose an algorithm which addresses the majority of the issues.
I will first present that algorithm, and then point out, as to why each concern is addressed.

\section{The BORING Algorithm}

We propose the following algorithm, called BORING. \textbf{BORING} stands for \textbf{B}ayesian \textbf{O}ptimization using \textbf{R}andom and \textbf{I}de\textbf{N}tifyable subspace \textbf{G}eneration.

The general idea of BORING can be captured in one formula, where $f$ stands for the real function that one wishes to approximate, and any subsequent function annotated by $g$ refers to a component of the right-hand side.

\begin{equation}
f(x) \approx g_0(A x) + \sum_{i \in \mathbb{Z}^+}^{q} g_i( A^{\bot} x )
\label{eq:dimRedEquation}
\end{equation} \\

Where the following variables have the following meaning
\begin{itemize}
\item $A$ is the active subspace projection (an element of the Stiefel manifold) learned through our algorithm, using Algorithm 1
\item $A^{\bot}$ is a matrix whose subspace is orthonormal to the projection of $A$.
We randomly generate $A^{\bot}$ using Algorithm 2.
\item The subscript $i$ in the right additive term denotes that we view each output dimension of $A^{\bot} X$ as independent to the other output dimensions.

\end{itemize}

I will now proceed with a more detailed description.

\subsection{Algorithm Description}

\subsubsection{Overview}

We explore a novel method which is based on additive GPs and an active subspace projection matrix.
We use different kinds of kernels.
We want to calculate $g_i$ and $A$ as defined in \ref{eq:dimRedEquation}, such that the log-likelihood of the data we have accumulated so far is maximized.\\
 
 The following few steps are applied after a "burn-in" period, in which we use random sampling to acquire new points.
 We sample the data points using UCB.
 This provides a set of data points which we can use to identify the subspace and the projection matrix.\\
 
 In simple terms, the algorithm proceeds as follows:
 
 \begin{enumerate}
 \item Pick the first $n$ samples using UCB sampling.
 During this period, we use UCB with a naive kernel, which has as many dimensions as the domains dimensions.
 From the collected points, approximate the active projection matrix $A$ using algorithm 1 from \citep{Tripathy}.
 \item Generate a basis that is orthonormal to every element in $A$.
 Concatenating these basis vectors $v_1, \ldots, v_{n-{q}}$ amongst the column-dimension gives us the passive projection matrix $A^\bot$.
 \item Maximize the GP for each individual expression of the space within $A$, and parallel to that also orthogonal to $A$ (as given by $A^\bot x$) individually. 
 \end{enumerate}
 
 This addresses the curse of dimensionality, as we can freely choose $q \geq d_e$ to set the complexity of the second term while the first term still allows for creating proximity amongst different vectors by projecting the vectors onto a smaller subspace.
The active subspace captures the direction of strongest direction, whereas the passive subspace projection captures an additional GP that adds to the robustness of the algorithm, should the subspace identification fail, or if the true function does not have a real subspace (i.e. is of the form $f(x) = f_0(Ax) + f_1(A^{\bot}x) \text{where} \lVert f_1 \rVert_{\infty} << \lVert f_0 \rVert_{\infty})$.
The additive terms that get projected onto the passive subspace also allow incorporating smaller perturbations in the space orthogonal to $A$ to occur.

\begin{algorithm}[H]
\caption{BORING Alg. 1 - Bayesian Optimization using BORING}

\begin{algorithmic} 
\STATE $X \leftarrow \emptyset$
\STATE $Y \leftarrow \emptyset$

\COMMENT{Burn in rate - don't look for a subspace for the first 100 samples}
\STATE $i \leftarrow 0$
\WHILE{i < 100}
\STATE $i++$
\STATE $x_* \leftarrow $ argmax$_x$ acquisitionFunction$(dot(Q^{\bot}, x) )$ using standard UCB over the domain of $X$.
\STATE Add $x_*$ to $X$ and $ f(x_*)$ to $Y$.
\ENDWHILE

\STATE $A, d, \phi \leftarrow $ Calculate active subspace projection using Algorithm 2 from the paper by Tripathy.
\STATE $A^{\bot} \leftarrow $ Generate passive subspace projection using Algorithm 3.
\STATE $Q \leftarrow $ colwiseConcat( $[A, A^{\bot}]$ ) 
\STATE $gp \leftarrow GP( $dot$( Q^T, X), Y)$
\STATE kernel $\leftarrow$ activeKernel + $\sum_i^{q}$ passiveKernel$_i$ 

\WHILE{we can choose a next point}
\STATE $x_* \leftarrow $ argmax$_x$ UCB$(dot(Q^{\bot}, x) )$
\STATE Add $x_*$ to $X$ and $ f(x_*)$ to $Y$.
\ENDWHILE

\RETURN $A, A^\bot$
\end{algorithmic}

\end{algorithm}

Where $\phi$ are the optimized kernel parameters for the activeKernel.
The active projection matrix using the following algorithm, which is identical to the procedure described in \citep{Tripathy}.
The generation of the matrix $ A^{\bot} $ is described next.

\subsubsection{Finding a basis for the passive subspace (a subspace orthogonal to the active subspace)}

\begin{equation}
A = 
\begin{bmatrix}
 \vdots & \vdots & & \vdots \\
 a_1 & a_2 & ... & a_{d_e} \\
 \vdots & \vdots & & \vdots
\end{bmatrix}
\label{eq:maximalEmbedding}
\end{equation}

Given that we choose a maximal lower dimensional embedding (maximizing the log-likelihood of the embedding for the given points), some other axes may be disregarded.
However, the axes that are disregarded may still carry information that can make search faster or more robust.

To enable a trade-off between time and search space, we propose the following mechanism.

Assume an embedding maximizing \ref{eq:maximalEmbedding} is found.
Then the active subspace is characterized by it's column vector $  a_1, a_2, ..., a_{d_e} $.
We refer to the space spanned by these vectors as the \textit{active subspace}.

However, we also want to address the subspace which is not addressed by the maximal embedding, which we will refer to \textit{passive subspace}.
This passive subspace is characterized by a set of vectors, that are pairwise orthogonal to all other column vectors in $A$, i.e., the vector space orthogonal to the active subspace spanned by the column vectors of $A$.

As such, we define the span of the active and passive subspace is defined by the matrix:

\begin{equation}
Q = 
\begin{bmatrix}
A & A^\bot
\end{bmatrix}
\label{eq:entireSubspace}
\end{equation}

where $A^\bot$ describes the matrix that is orthogonal to the column space of $A$.
For this, $A^\bot$ consists of any set of vectors that are orthogonal to all other vectors in $A$.\\

The vectors forming $A^\bot$  is generated by taking a random vector and applying Gram Schmidt.
This procedure is repeated for as many orthogonal vectors as we want. 
The procedure is summarised in Algorithm 3:\\

\begin{algorithm}[H]
\caption{BORING Alg. 3 - generate orthogonal matrix to A(A, n) }

\begin{algorithmic} 
\REQUIRE $A$ a matrix to which we want to create $A^{\bot}$ for; $n$, the number of vectors in $A^{\bot}$.

\STATE $Q \leftarrow$ emptyMatrix()
\COMMENT{ The final concatenated $Q$ will be $A^{\bot}$. }
\FOR{i = 1,...,n}
\STATE $i \leftarrow 0$ 
\WHILE{True}
\STATE i++
\STATE $q_i \leftarrow $ random vector with norm 1
\STATE newBasis = apply gram schmidt single vector( $[A, Q], q_i$ ) 

\IF{ dot(normed$A^T$, newBasis) $\approx \mathbf{0}$ \textbf{and} $\lVert$ newBasis $\rVert_2$ $> 1e-6$}
\STATE $Q \leftarrow$ colwiseConcatenate( $(Q, $ newBasis)
\STATE break
\ENDIF
\ENDWHILE                
\ENDFOR

\RETURN $Q$
\end{algorithmic}
\end{algorithm}

\subsubsection{Additive UCB acquisition function}

Because the function is decomposed into multiple additive components, the computation of the mean and variance needs to be adapted accordingly. 
Although I do not use this method in my experiments, it is still useful to mention one way to approximate these terms for higher dimensional input.
The following method proposed in \citep{Rolland}  is used. \\

\begin{align}
\mu_{t-1}^{(j)} &= k^j(x_*^{(j)}, X^j)\Delta^{-1}y \\
\left( \sigma_{t-1}^{(j)} \right)^2 &= k^j(x_*^{j}, x_*^{j}) - k^j(x_*^j, X^{(j)}) \Delta^{-1} k^j(X^{(j)}, x_*^j)
\end{align}

where $k(a, b)$ is the piecewise kernel operator for vectors or matrices $a$ and $b$ and $\Delta = k(X, X) + \eta I_n$.
A single GP with multiple kernels (where each kernel handles a different dimension of $dot(Q^T, x)$) is used.
There are $q+1$ kernels (the $+1$ comes from the first kernel being the kernel for the active subspace). \\

Using this information about each individual kernel component results in the simple additive mean and covariance functions, which can then be used for optimization by UCB:


\begin{align}
\mu(x) &= \sum_{i=1}^{M} \mu^{(i)} ( x^{(i)} ) \\
\kappa(x, x') &= \sum_{i=1}^{M} \kappa^{(i)} ( x^{(i)}, x'^{(i)}  )
\end{align}

One should notice that this additive acquisition function is an approximation of the real acquisition function. 
For lower dimensions - such as $d<3$ - it is not required to decompose the acquisition function into separate additive components.

\subsubsection{How does our algorithm address the shortcomings from chapter 4?}

\begin{enumerate}
\item Our algorithm intrinsically uses multiple restarts.
As such, bad initial states and bad projection matrices are discarded as better ones are identified.
This makes our algorithm more reliable than algorithms like naive REMBO (without interleavings).
\item Our algorithm allows to not only optimize on a given domain but also identify the subspace on which the maximal embedding is allocated on.
In addition to that, no gradient information is needed.
\item Our algorithm uses a "burn-in-rate" for the first few samples, which allows for efficient point search at the beginning, and later on switches to finding the actual, real subspace.
This means that we need to compute the embedding only once, and can then apply optimization on that domain.
Our algorithm allows for a comfortable choice of how much computation should be put into identifying the subspace.
\item Our algorithm is more accurate and robust, as we do not assume that there is a singular maximal subspace. 
We also take into consideration that there might be a perturbation on lower dimensions.
In that sense, our algorithm mimics the idea of projection pursuit \citep{ProjectionPursuit}, as it identifies multiple vectors to project to the lower subspace.
\end{enumerate}

\chapter{Evaluation}

\ifpdf
    \graphicspath{{07_Chapter6/Figs/Raster/}{07_Chapter6/Figs/PDF/}{07_Chapter6/Figs/}}
\else
    \graphicspath{{07_Chapter6/Figs/Vector/}{07_Chapter6/Figs/}}
\fi

\section{Evaluation Settings}

Appendix A presents a list of synthetic functions and real datasets that are used to evaluate the effectiveness of a Bayesian Optimization algorithm. 
I conduct experiments in the following settings as mentioned in chapter \ref{syntheticFunction}.
I must emphasize that I modified the algorithm of Tripathy et al., as it got stuck in local minima within the first five steps of algorithm 1.
For this, I set up the constraint that $\tau > 0$ (instead of $\tau \geq 0$). 
This allowed for small perturbations that kept on improving the loss.

\section{Quantitative evaluation}
To recapitulate, I will use log-likelihood, angle-difference measure and cumulative regret to compare the performance of different algorithms.
We present how the different algorithms perform on the regret measure using UCB as the acquisition function.
It is important to point out that all experiments capped the matrix identification step to about 30 minutes.
This is much less than in the original papers that we base the algorithm on.
The reason for this is that we want to have an acceptable comparison for medium-sized experiments, where time and computational resources can be restrictive (like on a users laptop). \\

I want to indicate whether the contribution of the performance comes from our subspace identification, or from the algorithm. 
For this, I start the discussion of every function with a plot that shows how the algorithm performs when the real subspace matrix $W_{\text{true}}$ is assumed to be found (Tripathy's algorithm is not applied; instead we return the $W_{\text{true}}$ instead of an approximated $\hat{W}$). \\

To keep the measurements fair across algorithms, I fix the noise variance of the GP and the kernel hyperparameters for each function.

I have multiple independent runs for each function.
However, as most of the runs show similar results, I display only one of them unless something is interesting to see.
The reader should notice that the individual runs do carry the same kernel parameters (unless an algorithm-specific function decides to change these).
This means that the algorithms should theoretically have similar properties as UCB on the vanilla function if the active subspace is identified or the dimensionality is effectively reduced. \\

In all of these graphs, we apply the subspace identification at the 100th timestep.
This means that we use the first 100 sampled points from UCB to identify a subspace.
Any other future projected point is projected onto this subspace before one optimization is taken.

Due to numerical errors recognized at later stages, the maximum dimension that I test over is 5.
When I set the real dimensionality of an environment (not the active dimensionality), the property $W^T \times W = I$ is violated.
Apart from that, all algorithms were implemented from scratch.
Any implemented gradients were unit-tested, and the respective analytical gradient was (successfully) compared with their numerical gradients.
To test the functionality of the subspace identification algorithm, I also wrote tests to make sure that the log-likelihood increases, and that some other properties presented in the respective paper are satisfied.

\subsection{Parabola}

The function to be learned and optimized over is the following:

\def\WParaboa2D{
\begin{bmatrix}
    0.500\\
    0.192
\end{bmatrix}}

\begin{equation}
f(x) = \left( \WParaboa2D^T x \right)^2
\end{equation}
where we have $x \in \mathbf{R}^2$ and $W \in \mathbf{R}^{2 \times 1}$.

\paragraph{Assume $\hat{W} = W_{\text{true}}$}: I present how the respective algorithms perform if we assume that Tripathy's Stiefel Manifold optimization finds the perfect matrix.
This measures how the algorithm performs when we assume perfect subspace identification.

\begin{figure}[H]
  \centering
      \includegraphics[width=0.5\textwidth]{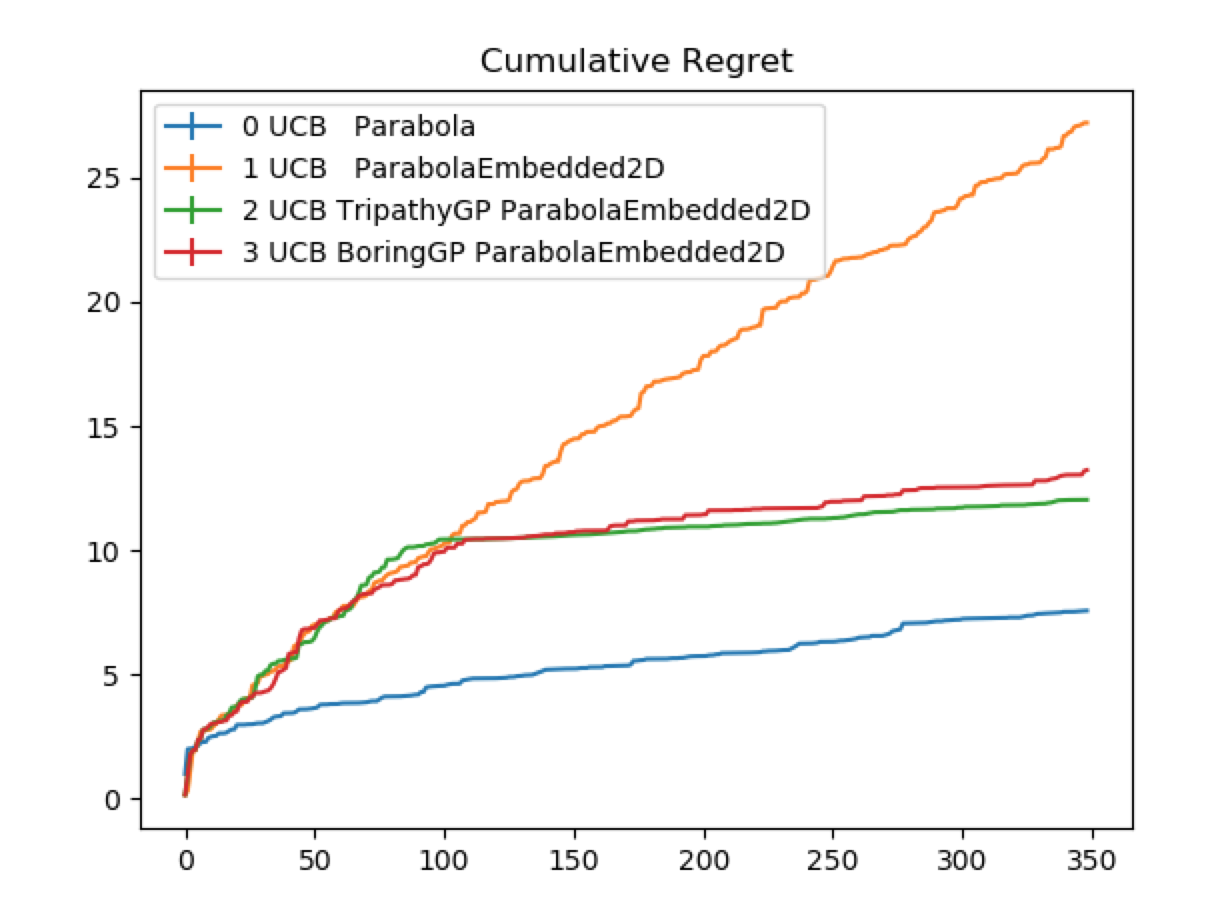}
  \caption{UCB on a Parabola embedded in 2D space, when we assume that tripathy's method finds the real projection matrix.}
\end{figure}

\paragraph{Assume $\hat{W} \neq W_{\text{true}}$}: I now proceed with how different algorithms perform on the function described above.
This measures how the algorithm performs, when subspace identification is a part of the optimization process.

\begin{figure}[H]
  \centering
      \includegraphics[width=0.5\textwidth]{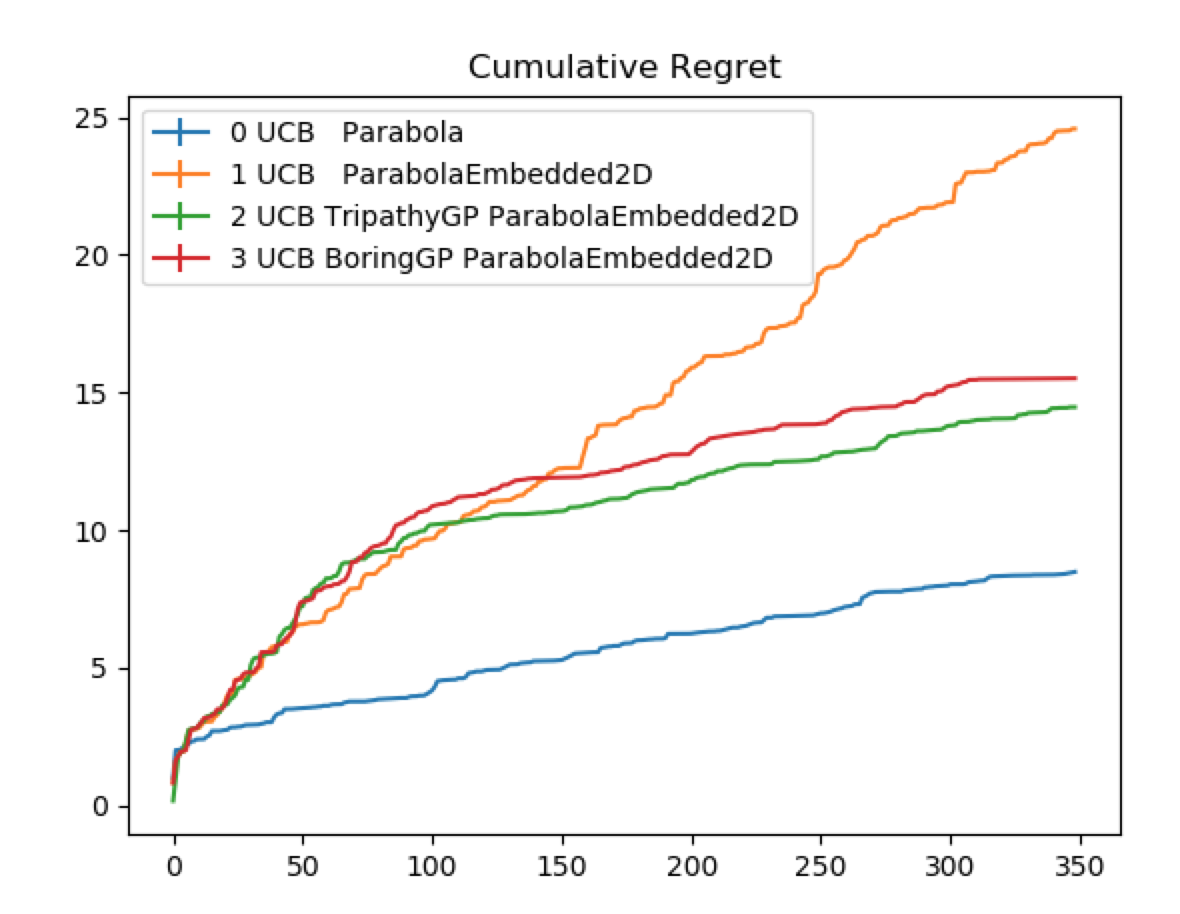}
  \caption{UCB on a Parabola embedded in 2D space.
  Tripathy's algorithm is applied to find a projection matrix $\hat{W}$.}
\end{figure}

One can easily see that the performance on UCB using Tripathy's matrix identification algorithm is similar to the case when we assume that Tripathy executes perfectly.
BORING is equivalent to Tripathy's performance, as the number of passive dimensions is set to 0.

\textbf{The log likelihood of the GP w.r.t the collected data points} of the Tripathy GP with the real matrix is comparable to the log-likelihood of the GP of the Tripathy model, where the active projection matrix is calculated using the algorithm (values of $-1.37$ and $-1.38$ or for a different run values of $195.32$ and $210.25$, where ranges are between  $-100$ and $700$).
One should notice, however, that the angle between the found matrix and the real projection matrix is almost always at $45°$ - a value that does not sound very intuitive, and for which the only reasonable explanation is that the optimization problem stays the same at this projection angle.
The reader can view graphs in a subsequent subsection.

\subsection{Camelback embedded in 3D}

The function to be learned and optimized over is the following:

\def\WCamelback3D{
\begin{bmatrix}
    -0.46554187 & -0.36224966 & 0.80749362 \\
     0.69737806 & -0.711918 & 0.08268378
\end{bmatrix}}

\begin{equation}
f(z_1, z_2) = \left( 4 - 2.1 * z_1^2 + \frac{z_1^4}{3} \right)  z_1^2 + z_1 *  z_2 + \left(-4 + 4 * z_2^2 \right) * z_2^2
\end{equation} \\

where we have $z = \left( \WCamelback3D^T x \right) $, and $ x \in \mathbf{R}^3$, $W \in \mathbf{R}^{2 \times 3}$ and where $z_1$ denotes the first entry of the $z$ vector, and $z_2$ denotes the second element of the $z$ vector.

\paragraph{Assume $\hat{W} = W_{\text{true}}$}: Again, I present how the respective algorithms perform if we assume that Tripathy's Stiefel Manifold optimization finds the perfect matrix.

\begin{figure}[H]
  \centering
      \includegraphics[width=0.5\textwidth]{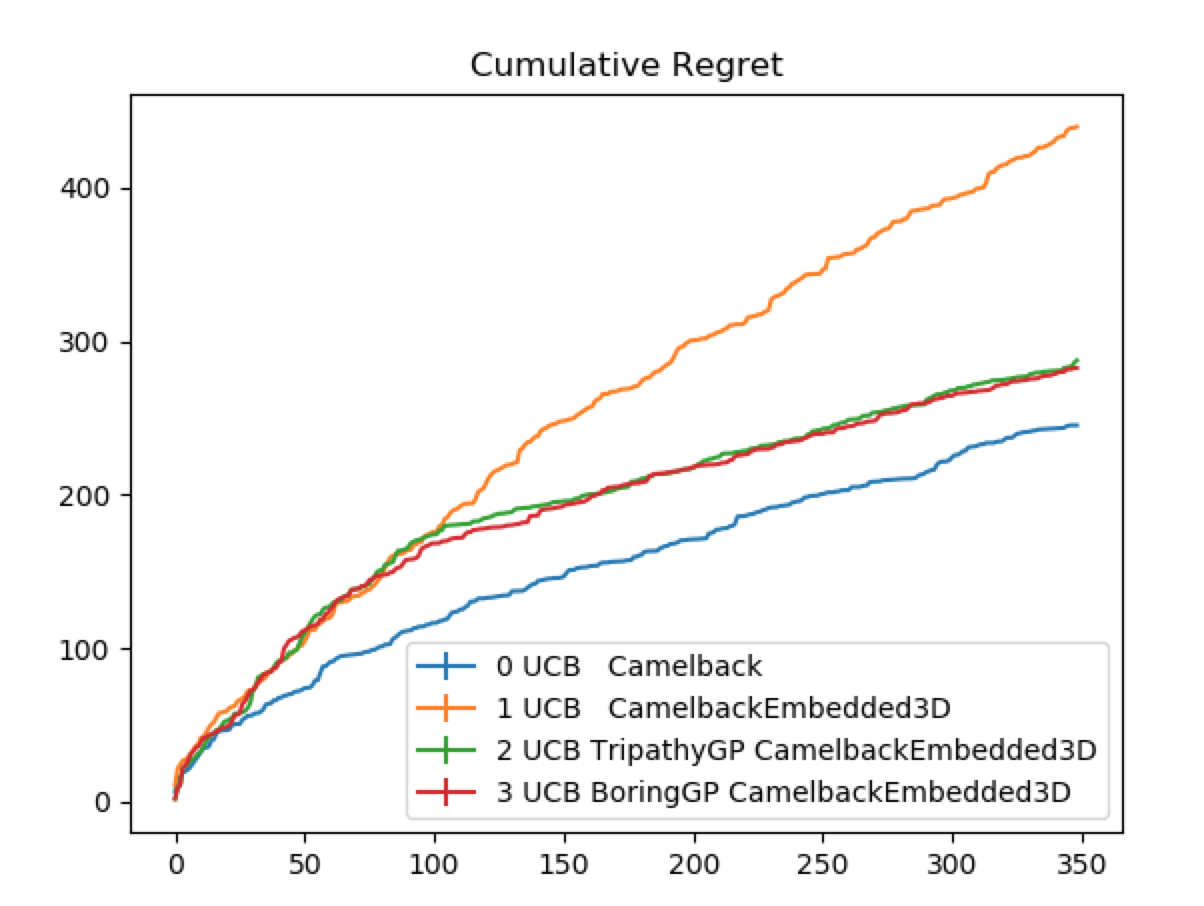}
  \caption{UCB on a 2D Camelback function embedded in 3D space.
  This is when we assume that Tripathy finds the real projection matrix $W_{\text{true}}$}
\end{figure}

\paragraph{Assume $\hat{W} \neq W_{\text{true}}$}: Again, I now proceed with the performance, when the subspace identification is part of the optimization process.
Again, BORING equals Tripthay's method, as setting the number of dimensions to 3 would make it use the entire search space (and thus not make it reduce the dimensionality).

\begin{figure}[H]
    \centering
    \begin{subfigure}[b]{0.40\textwidth}
        \includegraphics[width=\textwidth]{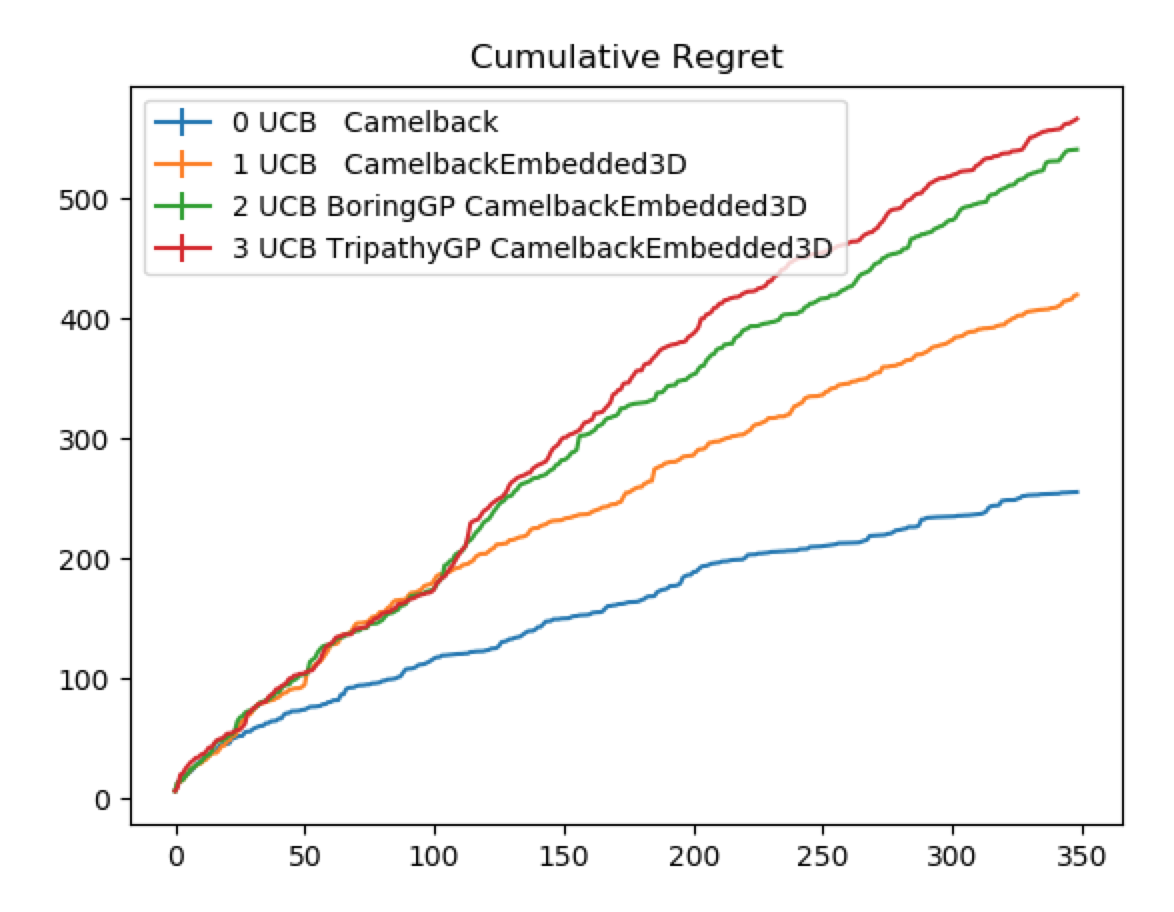}
        \label{fig:gull}
        \caption{Run 1}
    \end{subfigure}
    \begin{subfigure}[b]{0.40\textwidth}
        \includegraphics[width=\textwidth]{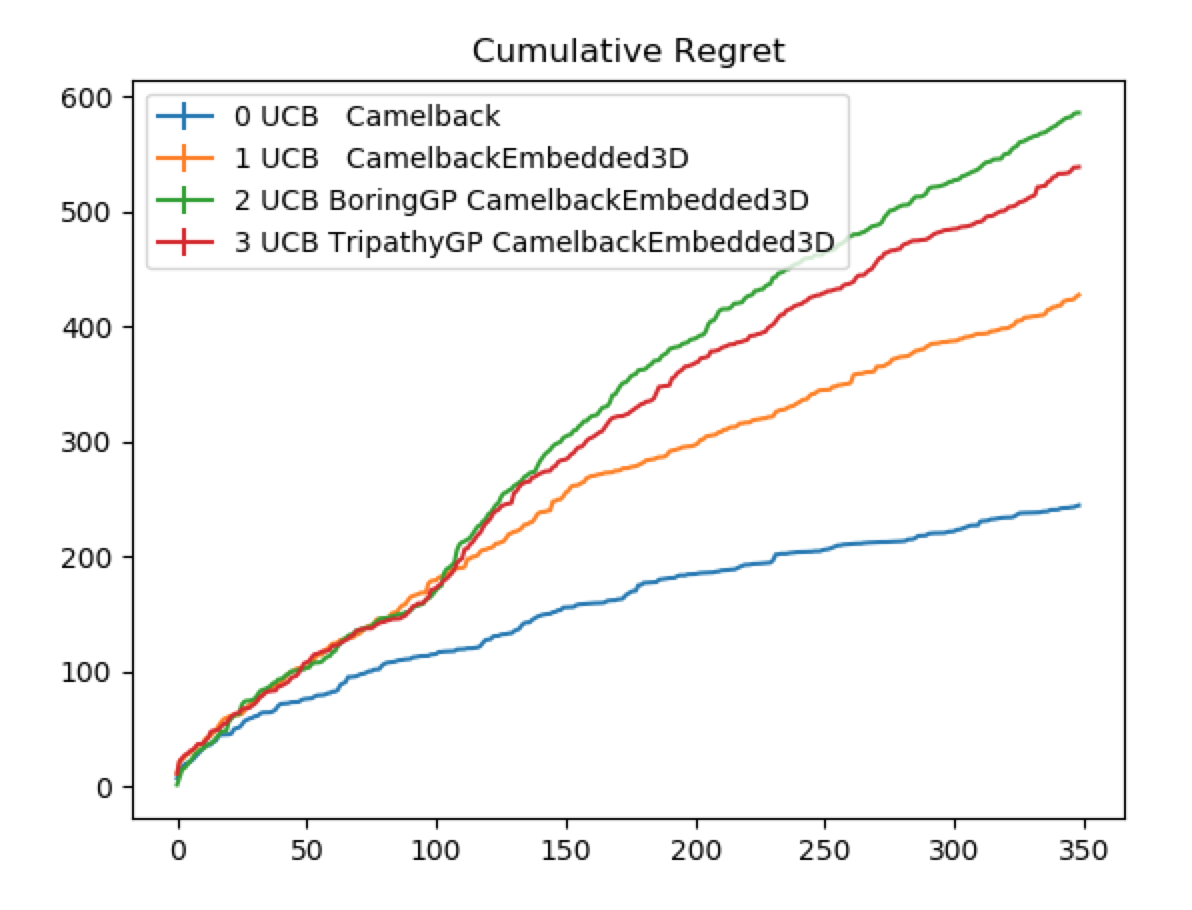}
        \label{fig:tiger}
        \caption{Run 2}
    \end{subfigure}   
        \caption{UCB on a 2D Camelback function embedded in 3D space.
          We apply Tripathy's algorithm to find a projection matrix $W$.}
\end{figure}

One can see that the subspace projection of Tripathy's method from 3D to 2D is not efficient. 
The difference between BORING and Tripathy is marginal, as they both rely on a similar algorithm.
This is an indication that the subspace projection is not close to the real subspace, but potentially finds a subspace which is acceptable when higher dimensions are taken into consideration.
To investigate this further, we increase the dimensionality of the domain in the next section.
Again, BORING uses only a 1D active subsapce, and has passive-dimensions 1 (i.e. the number of random embedding vectors is 1).
BORING performs similar to Tripathy's method.

\subsection{Camelback embedded in 5D}

\def\WCamelback5D{
\begin{bmatrix}
     -0.31894555 & 0.78400512 & 0.38970008 & 0.06119476 & 0.35776912 \\,
     -0.27150973 & 0.066002 & 0.42761931 & -0.32079484 &-0.79759551
\end{bmatrix}}

\begin{equation}
f(z_1, z_2) = \left( 4 - 2.1 * z_1^2 + \frac{z_1^4}{3} \right)  z_1^2 + z_1 *  z_2 + \left(-4 + 4 * z_2^2 \right) * z_2^2
\end{equation} \\

where we have \\
$z = \left( \WCamelback5D^T x \right) $, and $ x \in \mathbf{R}^5$, $W \in \mathbf{R}^{2 \times 5}$
and where $z_1$ denotes the first entry of the $z$ vector, and $z_2$ denotes the second element of the $z$ vector.

\paragraph{Assume $\hat{W} = W_{\text{true}}$}: The following shows how tripathy performs when we assume perfect projection-matrix identification.
BORING is identical to  Tripathy's method here.

\begin{figure}[H]
  \centering
      \includegraphics[width=0.5\textwidth]{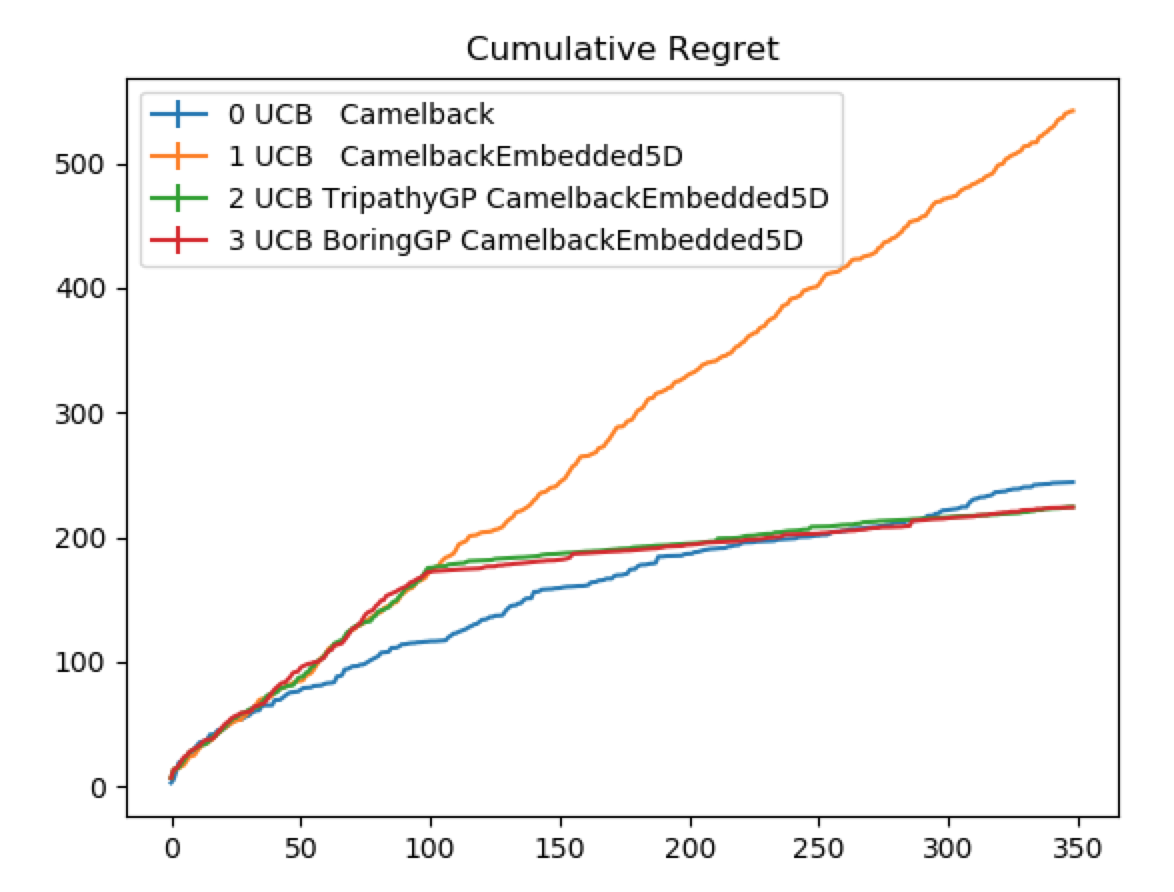}
  \caption{UCB on a 2D Camelback function embedded in 5D space.
  This is when we assume that Tripathy finds the real projection matrix $W_{\text{true}}$}
\end{figure}

The linear curve may be the result of kernel parameters that were not set very well, which may lead to the same point chosen repeatedly over and over again.
However, because the algorithm chooses these kernel parameters, I do not modify these to get square-root behaved UCB curves.

\paragraph{Assume $\hat{W} \neq W_{\text{true}}$}: The following curves present the performance of UCB when subspace identification is part of the optimization process.

\begin{figure}[H]
    \centering
    \begin{subfigure}[b]{0.40\textwidth}
        \includegraphics[width=\textwidth]{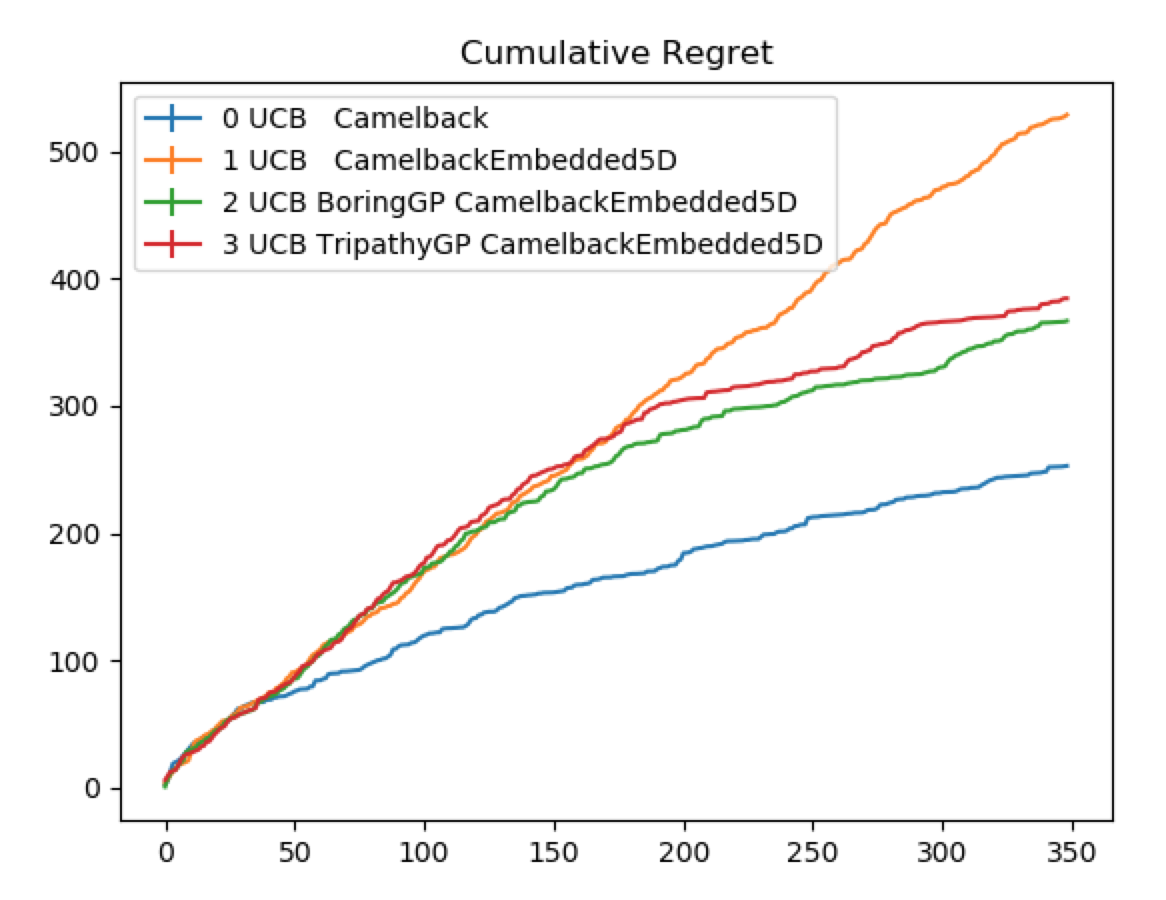}
        \label{fig:gull}
        \caption{Run 1}
    \end{subfigure}
    \begin{subfigure}[b]{0.40\textwidth}
        \includegraphics[width=\textwidth]{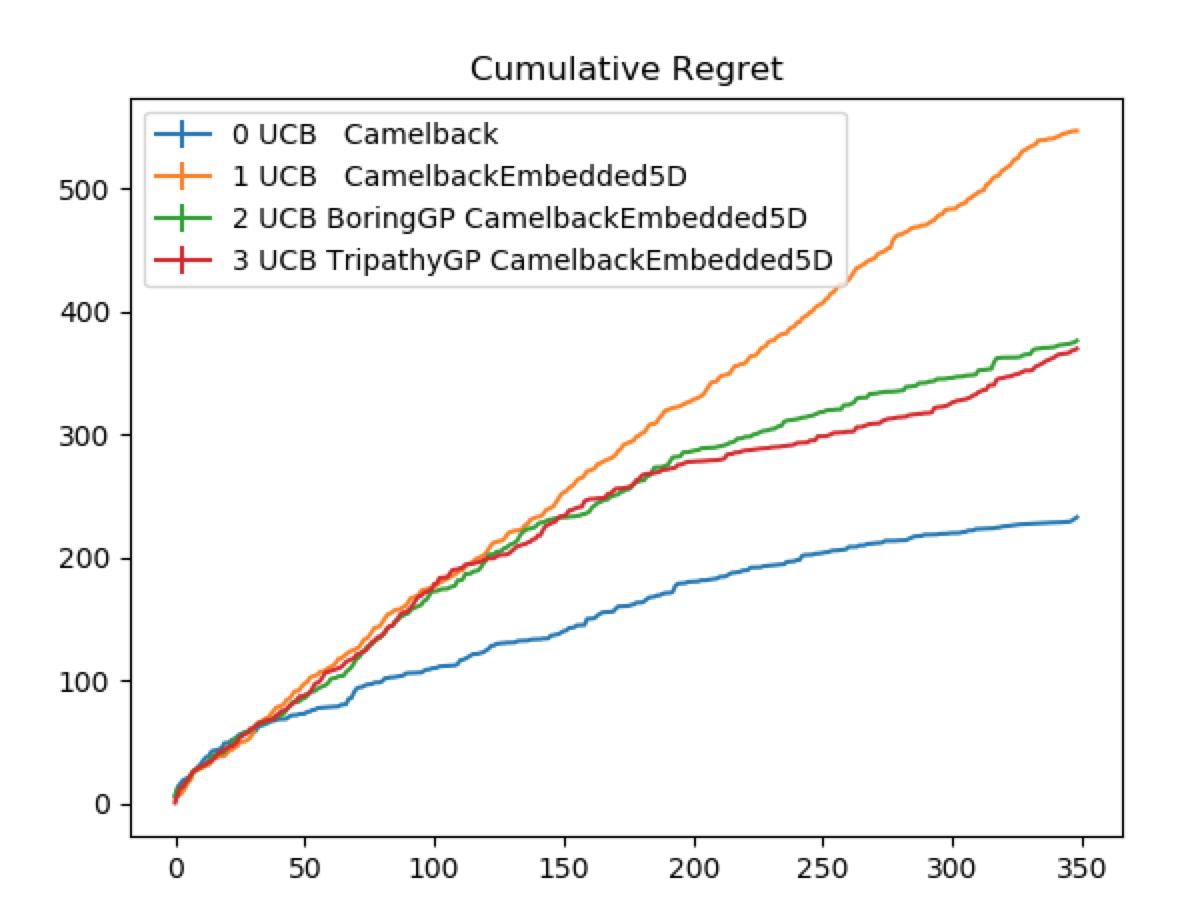}
        \label{fig:tiger}
        \caption{Run 2}
    \end{subfigure}   
           \caption{UCB on a 2D Camelback function embedded in 5D space.
  This is when we apply Tripathy's algorithm to find a projection matrix $W$, that is acceptable for optimization, but is not near close to the real projection matrix.}
\end{figure}

One can see for higher dimensions, Tripathy's method performs well, as it can reduce the dimensionality of the optimization problem.
However, one can see that the regret achieved from the empirical projection-matrix identification (the real projection-matrix is not found) is much higher.
This poses the question of how well the found projection matrix is compared to the real projection matrix.
BORING has a passive-dimension of 1 (the active dimensions is equal to Tripathy's number of active dimensions minus 1, which is equal to 1).
BORING achieves ok performance, as Tripathy's method is not able to identify the real projection matrix very well. 
I analyze the log-likelihood of the GP to get a quick answer. \\

\textbf{The log likelihood of the GP w.r.t the collected data points} of the Tripathy GP with the real matrix is not comparable to the log-likelihood of the GP of the Tripathy model, where the active projection matrix is calculated using the algorithm.
For different runs, the log-likelihood of the GP including the real matrix is at $-2$ and $-389$, whereas the log-likelihood of the GP with the estimated projection matrix is at $-0.92$ and $-102$. 
Although the values $-2$ and $-0.92$ are close to each other, the pair ($-389$, $-102$) shows that the subspace identification task for this algorithm is much more unstable than that for the parabola.
In later subsections, I will investigate this issue further.

\subsection{Exponential Sinusoidal}

The function to be learned and optimized over is the following:

\def\WSinusoidal5D{
\begin{bmatrix}
		-0.41108301 & 0.22853536 & -0.51593653 & -0.07373475 & -0.71214818 \\
           0.00412458 & -0.95147725 & -0.28612815 & -0.06316891 & -0.093885
\end{bmatrix}}

\begin{align}
f(z_1, z_2) & = f(z_1) + f(z_2) \\
& = \exp{\left[-\frac{1}{2} z_1 \right]}   \cos{(2 z_1)} + 0.01   \exp{\left[-\frac{1}{2} z_2 \right] }   \cos{(2 z_2)}
\end{align}

where we have \\
$z = \left( \WSinusoidal5D^T x \right) $, and $ x \in \mathbf{R}^5$, $W \in \mathbf{R}^{2 \times 5}$
and where $z_1$ denotes the first entry of the $z$ vector, and $z_2$ denotes the second element of the $z$.
The reader can notice that for the domain at hand (where $-3 < z < 3$), we have $ \lVert f_1 \rVert_{\infty} << \lVert f_0 \rVert_{\infty})$

\paragraph{Assume $\hat{W} = W_{\text{true}}$}: I present how the respective algorithms perform if we assume that Tripathy's Stiefel Manifold optimization finds the perfect matrix.
This measures how the algorithm performs when we assume perfect subspace identification.

\begin{figure}[H]
  \centering
      \includegraphics[width=0.5\textwidth]{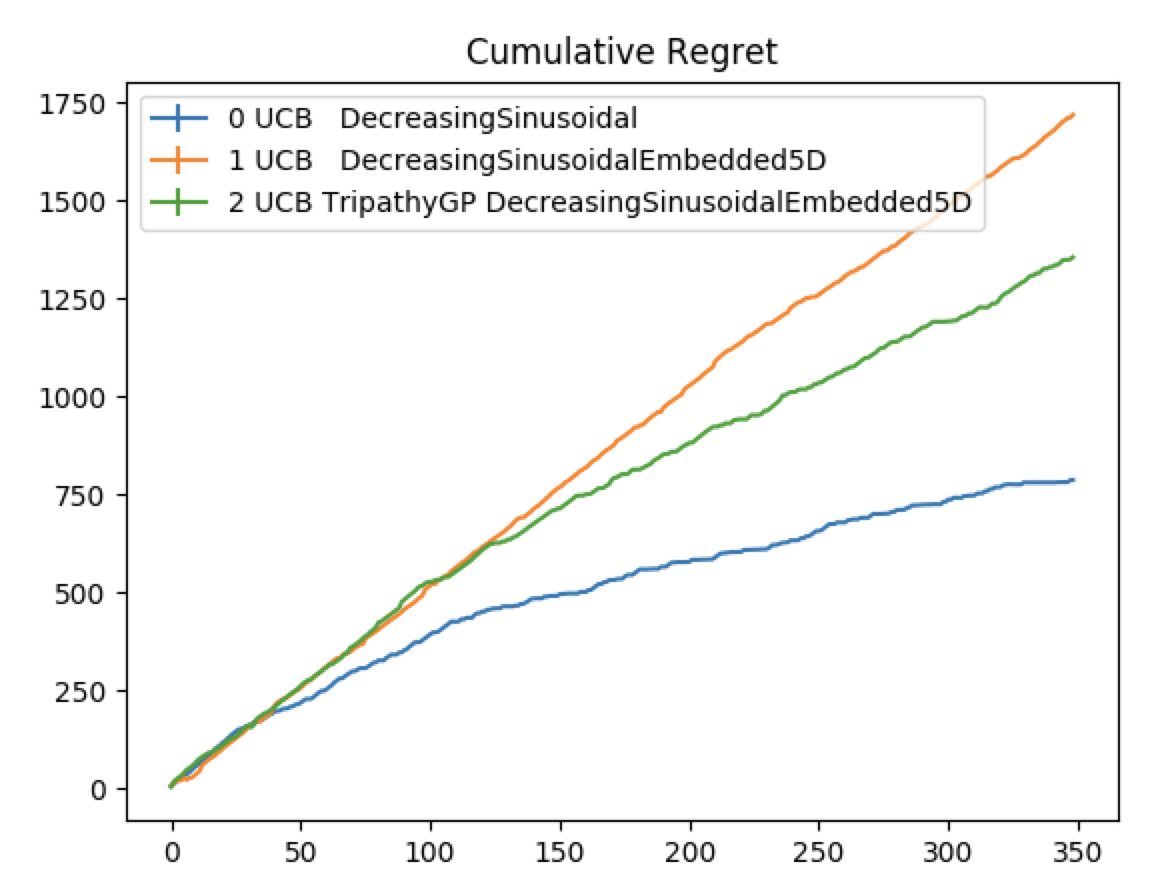}
  \caption{UCB on the Sinusoidal embedded in 2D space, when we assume that Tripathy's method finds the real projection matrix.
  One can see that the line is less curvy than then other examples.}
\end{figure}

\paragraph{Assume $\hat{W} \neq W_{\text{true}}$ and $d=1$}: I now proceed with how different algorithms perform on the function described above.
This measures how the algorithm performs, when subspace identification is a part of the optimization process.

\begin{figure}[H]
	\centering
    \begin{subfigure}[b]{0.5\textwidth}
        \includegraphics[width=\textwidth]{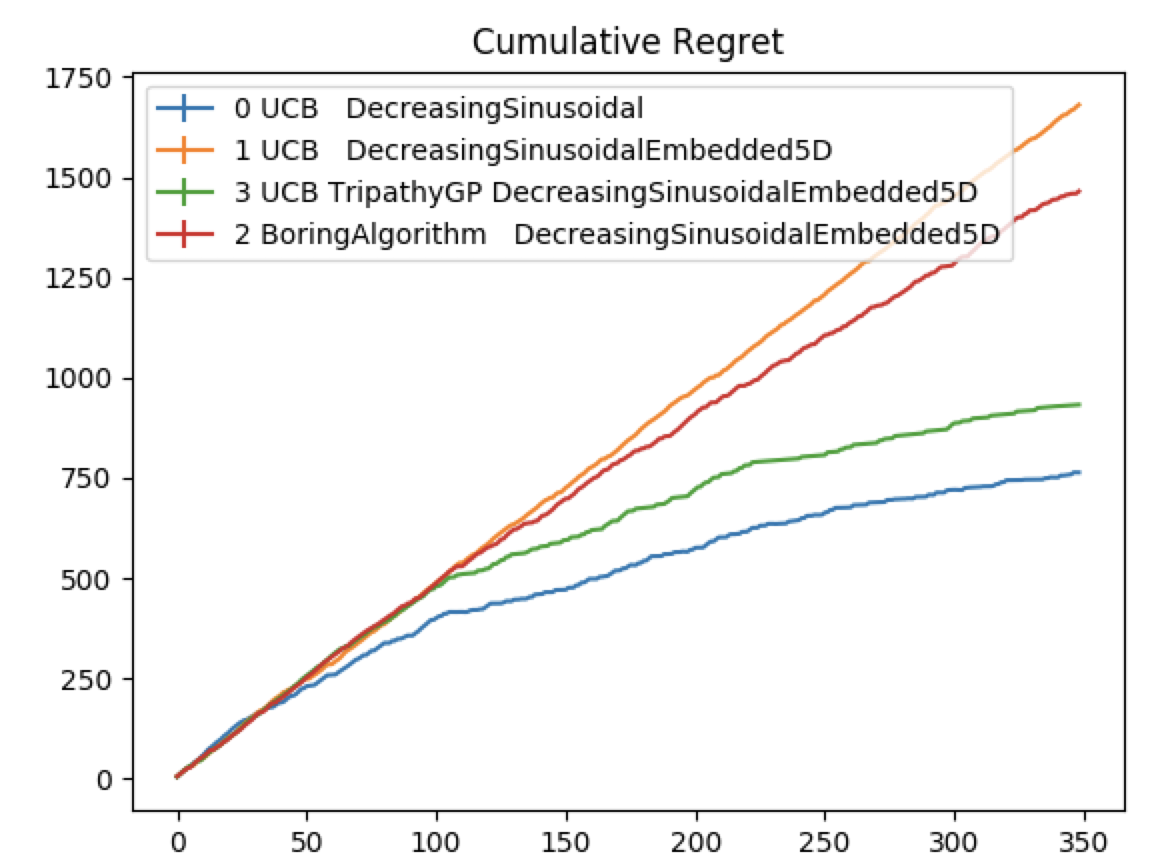}
        \label{fig:tiger}
        \caption{Run 2}
    \end{subfigure}   
           \caption{UCB on the 2D Sinusoidal function embedded in 5D space.
  This is when we apply Tripathy's algorithm to find a projection matrix $W$, that is acceptable for optimization, but is not near close to the real projection matrix.}
\end{figure}

One can see the performance of Tripathy's method is superior to BORING.
The reason for this may be one of following:

\begin{enumerate}
\item Hyperparameters are not optimized as part of the process, when applying the BORING method (according to the new projection matrix of BORING $Q$).
\item BORING adds another dimension that the optimizer must take into consideration.
The overhead of this additional dimension is higher than the potential gain that BORING could exploit by looking at the smaller perturbations.
\end{enumerate}

To test the first assumption, I optimize the GP hyperparameters after the projection matrix (including the passive projection vectors) has been identified and generated.
This yielded no recognizable improvement to the above curves.
I was not able to test the second assumption, as my implementation of Tripathy's method was not stable for dimensions bigger than 5, and as this would require a significant amount of time to fix.
Thus, is it still an open question if it is advantageous to use BORING instead of Tripathy's method for higher dimensions.
The reason why BORING and Tripathy's method perform similary in Camelback can be seen in the next subsection, where we show how well the subspace identification performs from a log-likelihood, and an angle-difference perspective.

\textbf{The log likelihood of the GP w.r.t the collected data points} of the Tripathy GP with the real matrix is comparable to the log-likelihood of the GP of the Tripathy model, where the active projection matrix is calculated using the algorithm (values of $-1.37$ and $-1.38$ or for a different run values of $195.32$ and $210.25$, where ranges are between  $-100$ and $700$).
One should notice, however, that the angle between the found matrix and the real projection matrix is almost always at $45°$ - a value that does not sound very intuitive, and for which the only reasonable explanation is that the optimization problem stays the same at this projection angle.
The reader can view graphs in a subsequent subsection.

\paragraph{Assume $\hat{W} \neq W_{\text{true}}$ and active d=$2$ }: For a quick sanity check, I now investigate how Tripathy's method performs when we set the active number of dimensions to 2.

\begin{figure}[H]
  \centering
      \includegraphics[width=0.5\textwidth]{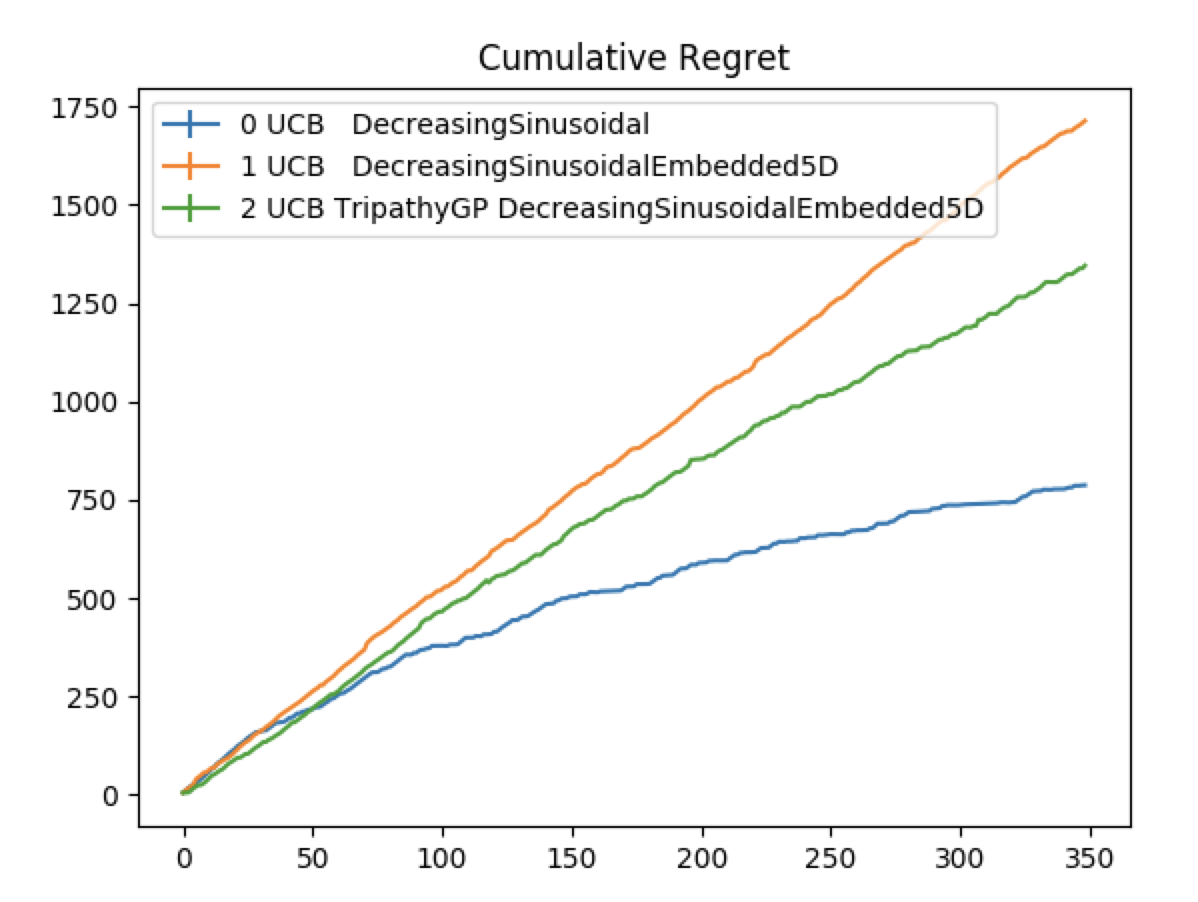}
  \caption{UCB on the 2D Sinusoidal embedded in 5D space, when we assume that Tripathy's method is applied to approximate the real projection matrix.
  One can see good performance in subspace recovery.}
\end{figure}

At this point the reader may wonder why adding random orthogonal vectors to the orthogonal projection matrix is necessary, if one can simply increase the number of dimensions for Tripathy's method.
The answer to this is that each individual small perturbation term would need to have an individual vector added to the projection matrix.
There number of vectors added to the projection matrix would be linear to the axis of small perturbations.
By adding a random vector, we account for a multitude of such functional-components, without adding linearly many vectors to the projection matrix.

\subsection{Log-Likelihood and Angle difference measures}

An interesting quantity to take into consideration is the log-likelihood of the sampled data with respect to the GP, and the angle between the found projection matrix, and the real projection matrix.
In the following, I describe how these quantities change over a function of time.
More specifically, the time refers to the number of steps that I allow for Tripathy's method to optimize over these parameters.

\paragraph{Parabola}

From the graphs, we can see that the parabola always seems to converge at a projection that is at a 45° to the real projection matrix. 
Intuitively, this seems odd.
One should notice, that the maximum of the optimization problem is the same when the space is rotated by 45°.
Another explanation could be that 100 data points on a 2D space are numerous enough, such that any matrix that does not directly map to the nullspace of the real projection matrix is an acceptable matrix.
Why the algorithm always converges at a matrix at 45° to the real projection matrix, would be unclear however in this case.

\begin{figure}[H]
    \centering
    \begin{subfigure}[b]{0.4\textwidth}
        \includegraphics[width=\textwidth]{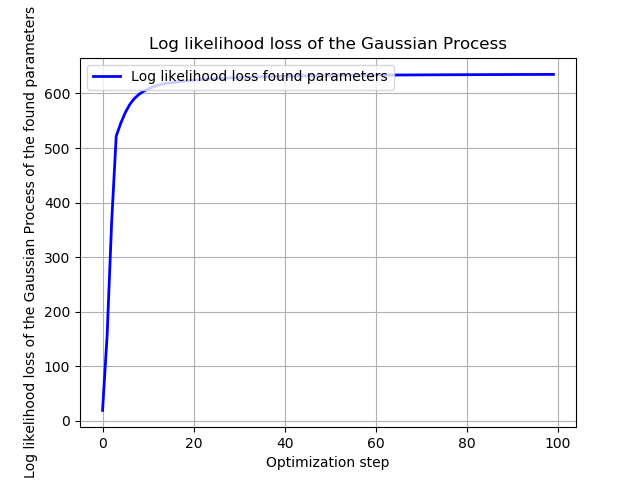}
        \label{fig:gull}
           \caption{The momentary $W$ from tripathy's algorithm, and it's log-likelihood w.r.t. the sampled data}
    \end{subfigure}
    \quad
        \begin{subfigure}[b]{0.40\textwidth}
        \includegraphics[width=\textwidth]{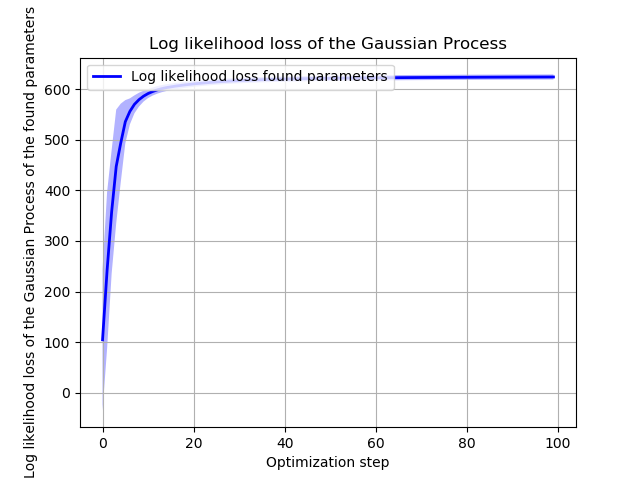}
        \label{fig:gull}
           \caption{The average of all momentary $W$ from Tripathy's algorithm, and it's log-likelihood to the sampled data}
    \end{subfigure}    \vskip\baselineskip
        \begin{subfigure}[b]{0.40\textwidth}
        \includegraphics[width=\textwidth]{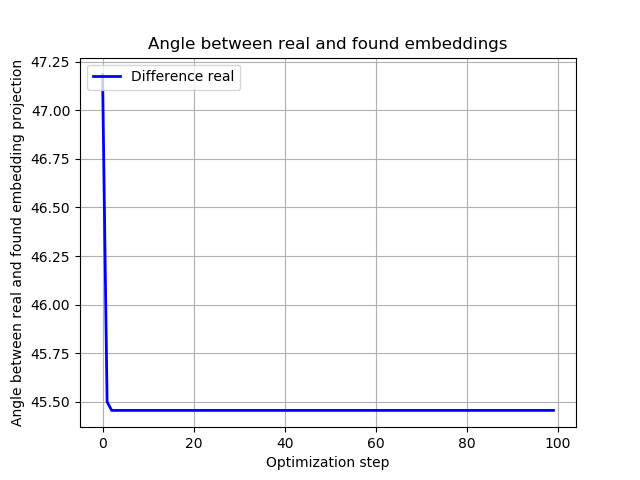}
        \label{fig:gull}
           \caption{The momentary $W$ from Tripathy's algorithm, and its angle to the real projection matrix}
    \end{subfigure}
        \quad
        \begin{subfigure}[b]{0.40\textwidth}
        \includegraphics[width=\textwidth]{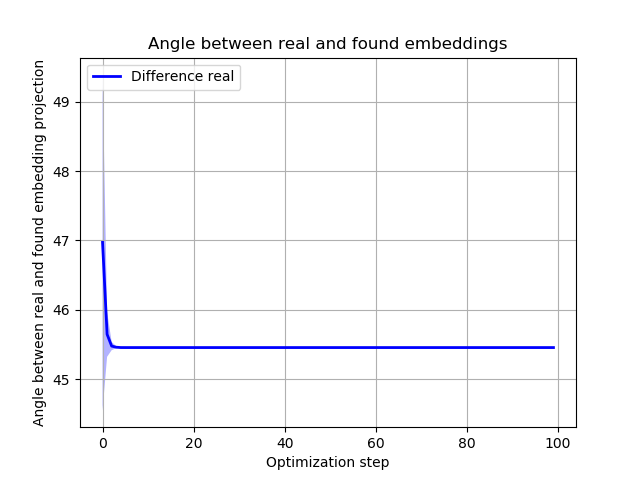}
        \label{fig:gull}
           \caption{The average of all momentary $W$ from Tripathy's algorithm, and it's angle to the real projection matrix}
          \end{subfigure}
    \caption{Log-Likelihood (top) and Angle (bottom) performance measures for a 1D Parabola embedded in a 2D space.
    The left graphs show the values for the run that was chosen as the "found" projection matrix.
    The right graphs show the average values over all restarts of tripathy's method.
    }\label{fig:animals}
\end{figure}

\paragraph{Camelback}
Because from the UCB experiments, we assume Camelback to be more unstable, we show the results of two independent runs that exhibit different behavior.
This is evidence, which Tripathy's algorithm on Camelback does not run stable, and has high variance (i.e., is not as robust).

\begin{figure}[H]
    \centering
    \begin{subfigure}[b]{0.40\textwidth}
        \includegraphics[width=\textwidth]{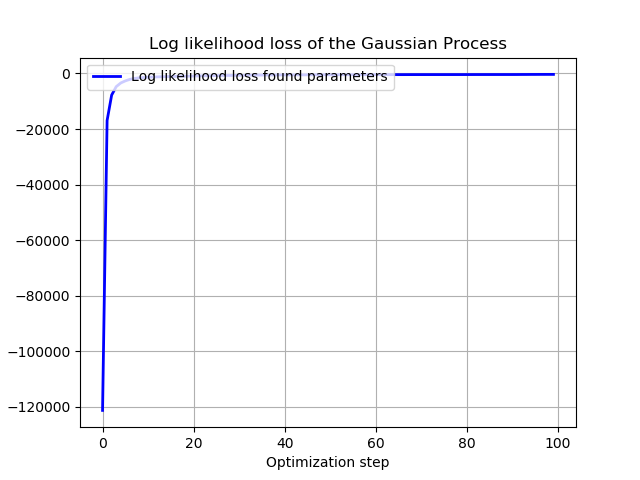}
        \label{fig:gull}
                           \caption{The momentary $W$ from tripathy's algorithm, and it's log-likelihood w.r.t the sampled data}
    \end{subfigure}
       \quad
   \begin{subfigure}[b]{0.40\textwidth}
        \includegraphics[width=\textwidth]{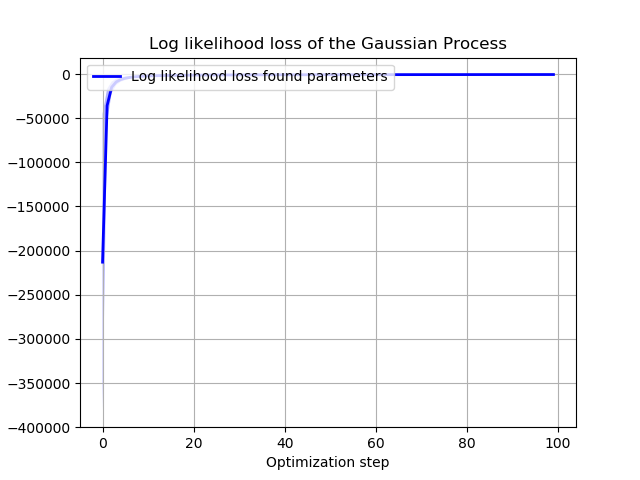}
        \label{fig:gull}
                           \caption{The average of all momentary $W$ from tripathy's algorithm, and it's log-likelihood w.r.t the sampled data}
    \end{subfigure} 
 \vskip\baselineskip
        \centering
    \begin{subfigure}[b]{0.40\textwidth}
        \includegraphics[width=\textwidth]{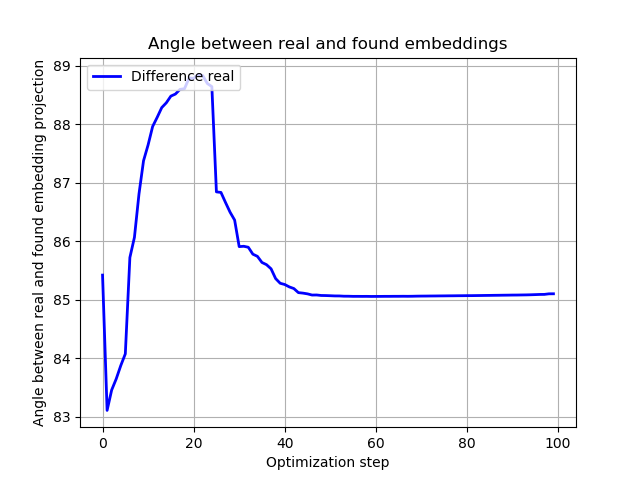}
        \label{fig:gull}
               \caption{The momentary $W$ from Tripathy's algorithm, and its angle to the real projection matrix}
    \end{subfigure}
        \quad
        \begin{subfigure}[b]{0.40\textwidth}
        \includegraphics[width=\textwidth]{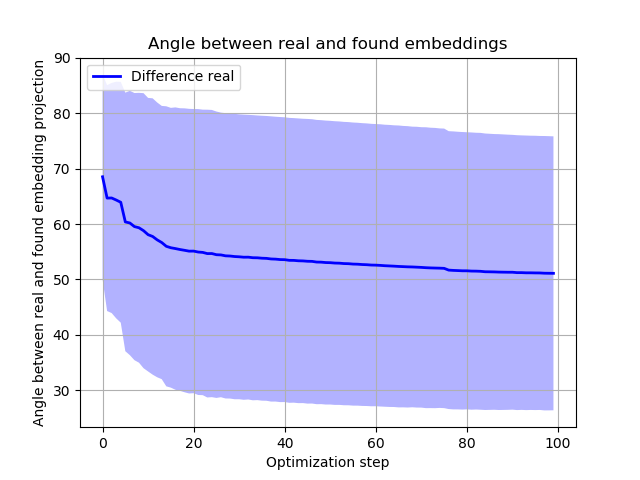}
        \label{fig:gull}
                           \caption{The average of all momentary $W$ from Tripathy's algorithm, and it's angle to the real projection matrix}
    \end{subfigure}
    \caption{Log-Likelihood (top) and Angle (bottom) performance measures for a 2D Camelback embedded in a 5D space.
    The left graphs show the values for the run that was chosen as the "found" projection matrix.
    The right graphs show the average values over all restarts of Tripathy's method.
    These are the results for run 1.
    }\label{fig:animals}
\end{figure}

\begin{figure}[H]
\center
    \begin{subfigure}[b]{0.40\textwidth}
        \includegraphics[width=\textwidth]{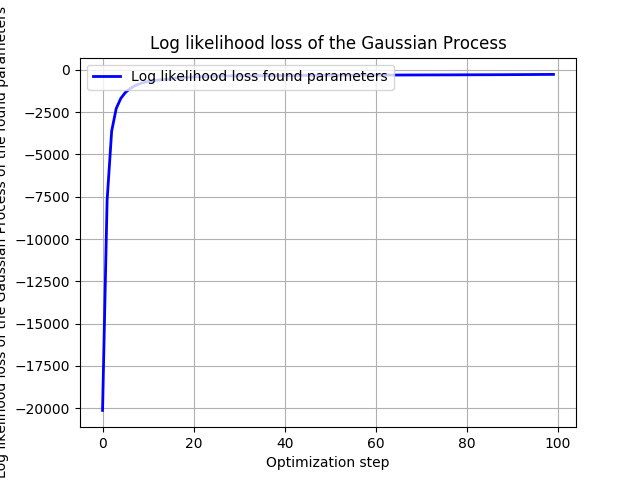}
        \label{fig:tiger}                          
         \caption{The momentary $W$ from tripathy's algorithm, and it's log-likelihood w.r.t the sampled data}
    \end{subfigure}   
        \quad
    \begin{subfigure}[b]{0.40\textwidth}
        \includegraphics[width=\textwidth]{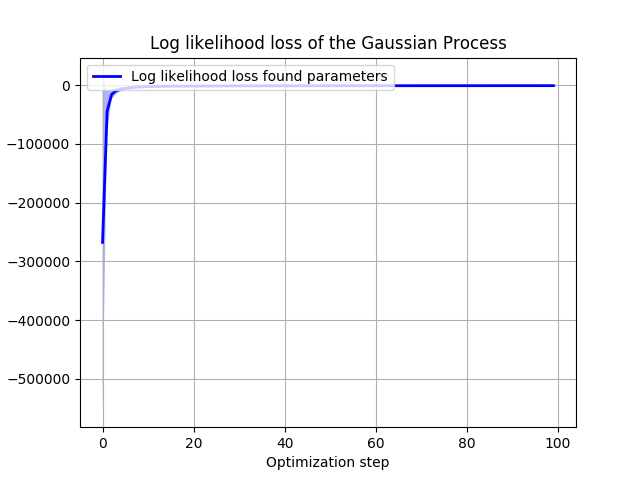}
        \label{fig:tiger}
               \caption{The momentary $W$ from tripathy's algorithm, and it's log-likelihood w.r.t. to the GP and the so-far collected data}
    \end{subfigure}   
 \vskip\baselineskip
         \centering
             \begin{subfigure}[b]{0.40\textwidth}
        \includegraphics[width=\textwidth]{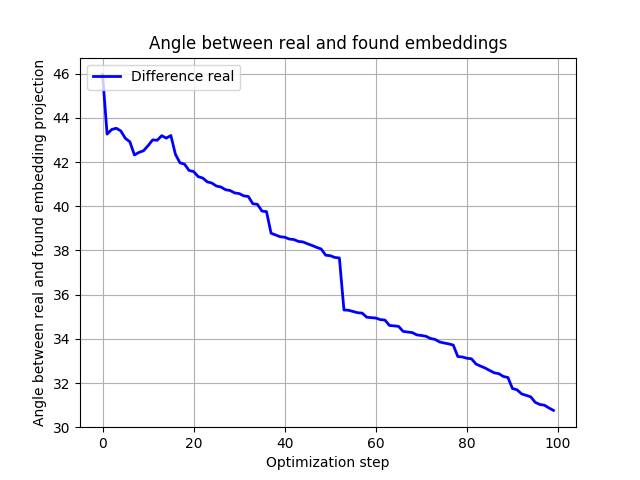}
        \label{fig:tiger}                           
        \caption{The average of all momentary $W$ from tripathy's algorithm, and it's log-likelihood w.r.t the sampled data}
    \end{subfigure}
        \quad
    \begin{subfigure}[b]{0.40\textwidth}
        \includegraphics[width=\textwidth]{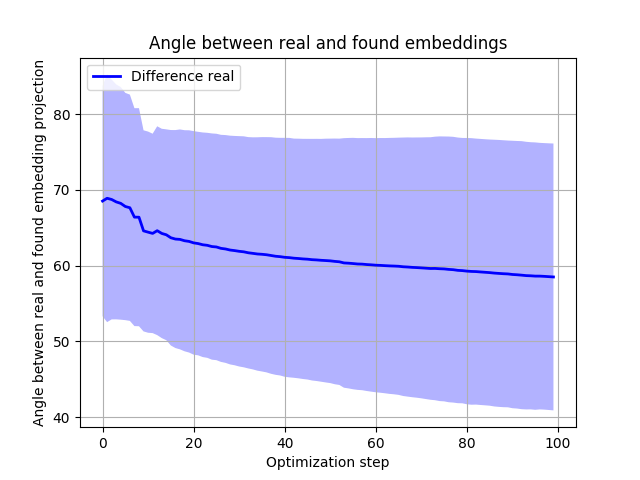}
        \label{fig:tiger}
               \caption{The average of all momentary $W$ from Tripathy's algorithm, and it's angle to the real projection matrix}
    \end{subfigure}   
        \caption{Log-Likelihood (top) and Angle (bottom) performance measures for a 2D Camelback embedded in a 5D space.
    The left graphs show the values for the run that was chosen as the "found" projection matrix.
    The right graphs show the average values over all restarts of Tripathy's method.
    These are the results for run 2.
    }\label{fig:animals}
\end{figure}

\paragraph{Sinusoidal}: 
Although we do not show the UCB curves for the sinusoidal (which also prove to be successful, when this 2d function is hidden within a 5D space), good results can be obtained for subspace identification.

\begin{figure}[H]
\center
    \begin{subfigure}[b]{0.40\textwidth}
        \includegraphics[width=\textwidth]{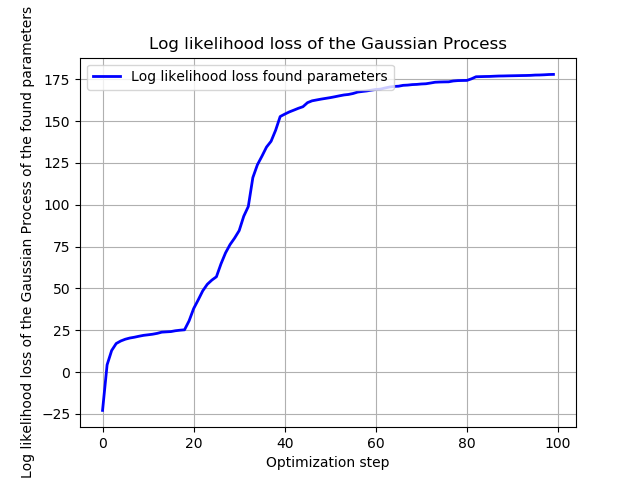}
        \label{fig:gull}
         \caption{The momentary $W$ from tripathy's algorithm, and it's log-likelihood w.r.t the sampled data}
    \end{subfigure}
       \quad
        \begin{subfigure}[b]{0.40\textwidth}
        \includegraphics[width=\textwidth]{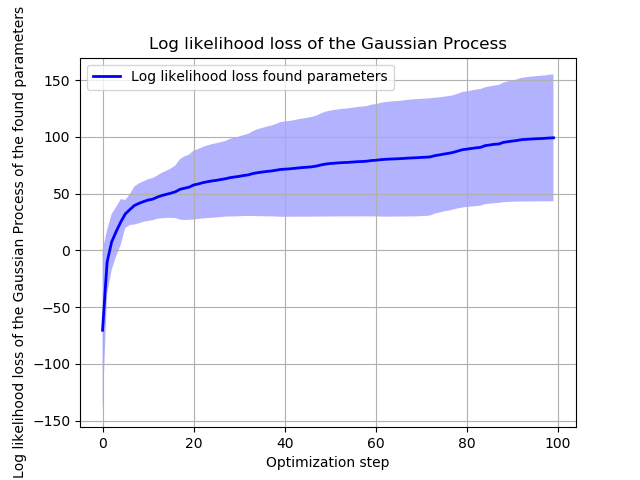}
        \label{fig:gull}
        \caption{The average of all momentary $W$ from tripathy's algorithm, and it's log-likelihood w.r.t the sampled data}
    \end{subfigure}

 \vskip\baselineskip
         \centering           

    \begin{subfigure}[b]{0.40\textwidth}
        \includegraphics[width=\textwidth]{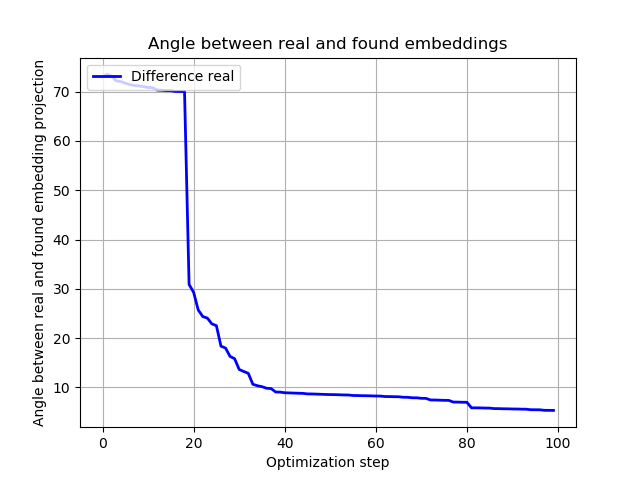}
        \label{fig:gull}
               \caption{The momentary $W$ from Tripathy's algorithm, and its angle to the real projection matrix}
    \end{subfigure}
        \quad
    \begin{subfigure}[b]{0.40\textwidth}
        \includegraphics[width=\textwidth]{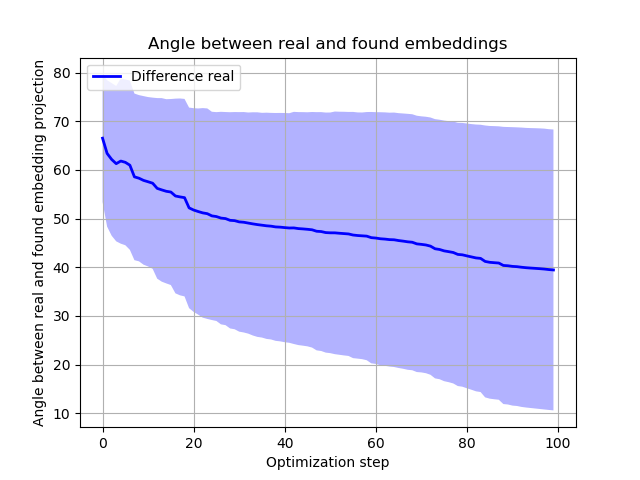}
        \label{fig:gull}
               \caption{The average of all momentary $W$ from Tripathy's algorithm, and it's angle to the real projection matrix}
    \end{subfigure}
           
        \caption{Log-Likelihood (top) and Angle (bottom) performance measures for a 2D Sinusoidal function embedded in a 5D space.
    The left graphs show the values for the run that was chosen as the "found" projection matrix.
    The right graphs show the average values over all restarts of Tripathy's method.
    These are the results for run 2.
    }\label{fig:animals}
\end{figure}

\section{REMBO}

The final algorithm I am investigating is REMBO, as described in chapter 2.
REMBO generally finds proper embeddings under the assumption, that optimization domain is normalized, and scaled to $[-\sqrt{d}, \sqrt{d}]^d$ where $d$ is at least the effective dimensionality of the optimization function at hand. 
The reader should acknowledge that the algorithm that was used to create the below plots only allowed to sample orthogonal matrices (instead of random matrices in REMBO).
This is a detail I noticed too late and could not change timely.
However, running two experiments for each environment with the randomly sampled matrix (without the orthogonality constraint) yielded similar results with a higher number of bad projections.
Because the results were similar besides the higher number of bad projections, I decided to keep the plots including the orthogonally sampled matrix. \\

I shortly present results for UCB that use REMBO as their optimization algorithm.
The high probability of failing implies a high variance amongst runs. 
As such, I perform three runs for each function addressed in the above section.

\paragraph{Parabola}
As one can see, REMBO usually finds an embedding that is acceptable and accelerates the optimization process.
However, run two shows that it can also fail in the simplest case of the parabola.
In this regard, Tripathy's method is more robust, as it does use some heuristic to check if the chosen matrix delivers good results.

\begin{figure}[H]
\center
    \begin{subfigure}[b]{0.30\textwidth}
        \includegraphics[width=\textwidth]{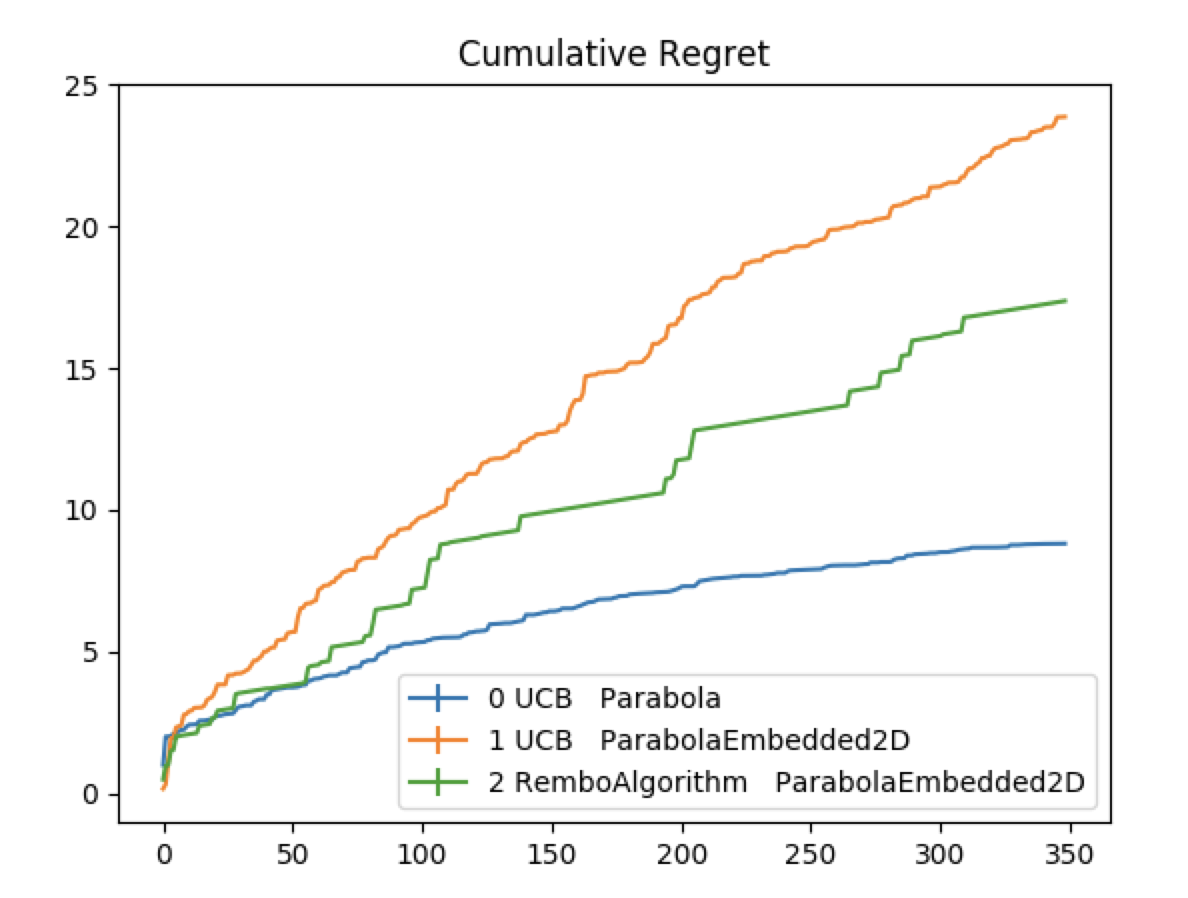}
        \label{fig:gull}
         \caption{Run 1}
    \end{subfigure}
        \begin{subfigure}[b]{0.30\textwidth}
        \includegraphics[width=\textwidth]{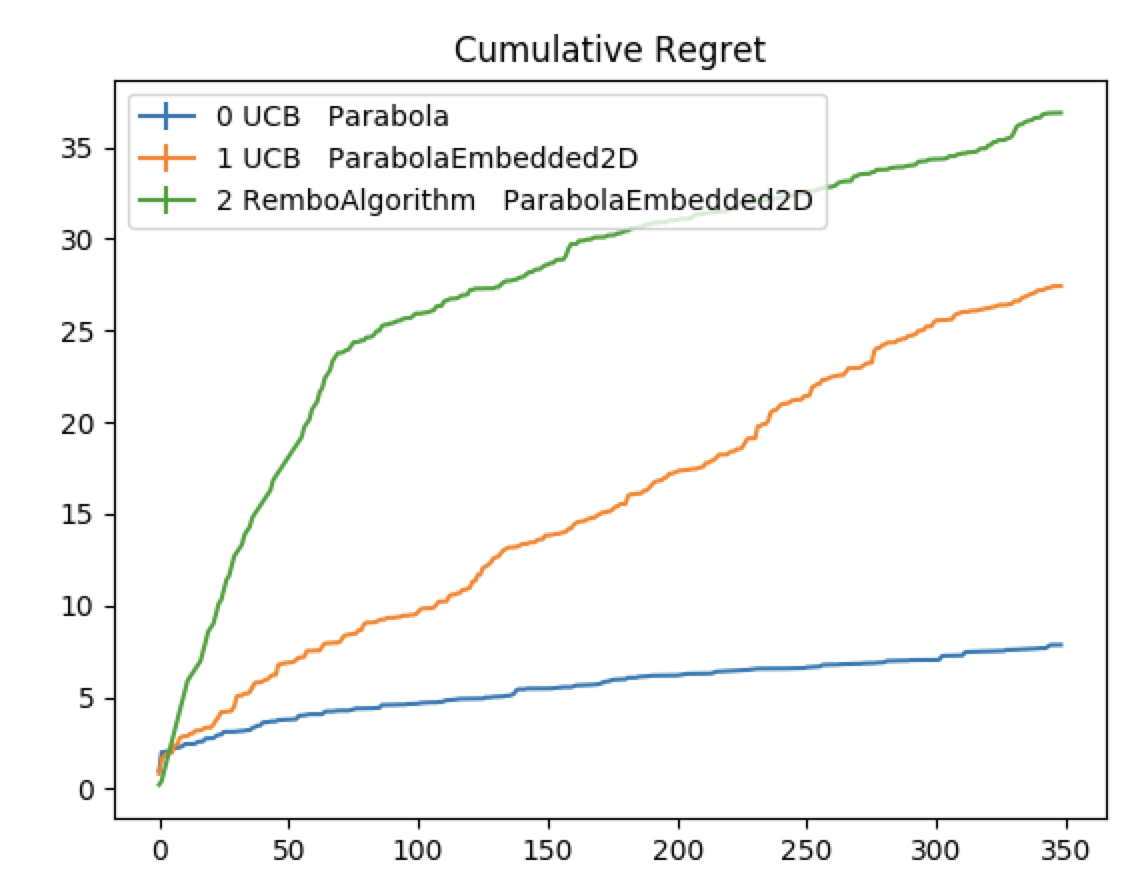}
        \label{fig:gull}
        \caption{Run 2}
    \end{subfigure}
    \begin{subfigure}[b]{0.30\textwidth}
        \includegraphics[width=\textwidth]{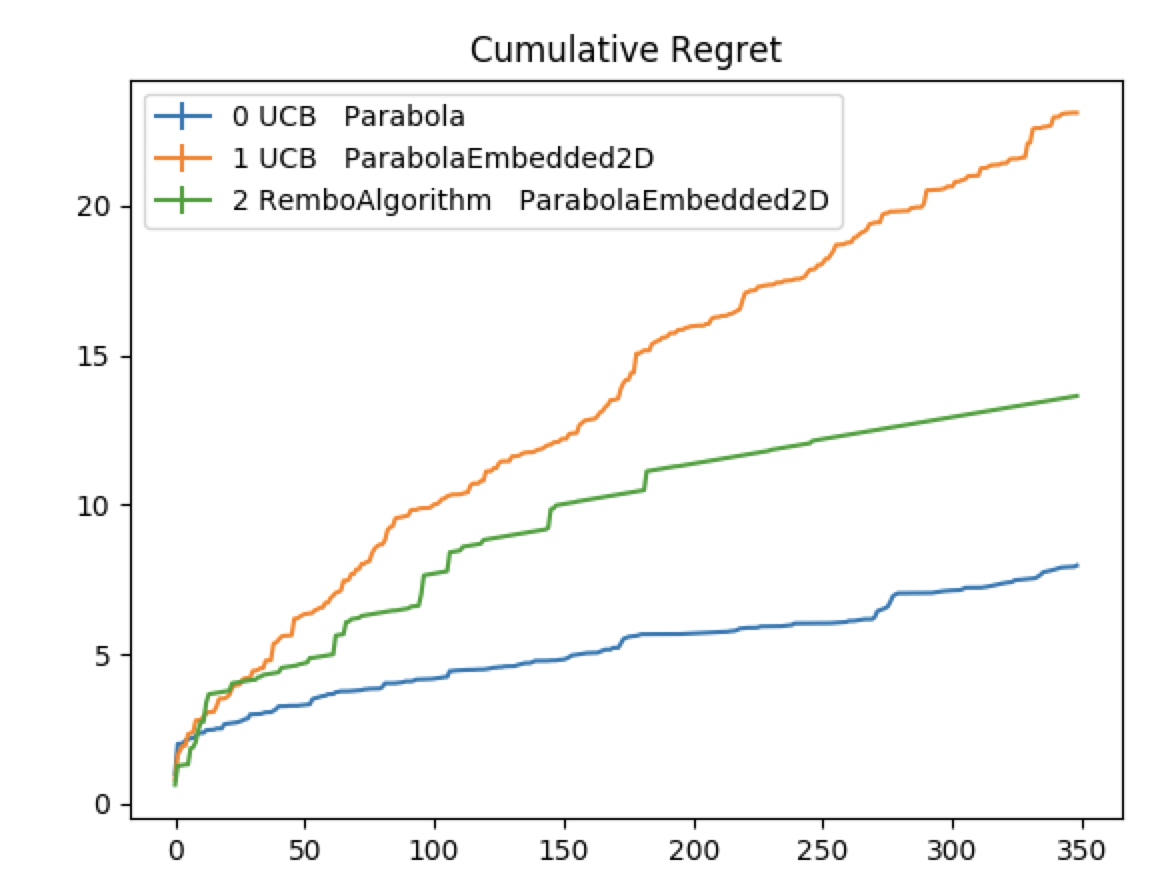}
        \label{fig:gull}
               \caption{Run 3}
    \end{subfigure}
        \caption{UCB using REMBO on a 1D Parabola embedded in a 2D space.
    }\label{fig:animals}
\end{figure}

\paragraph{Camelback3D}
This example illustrates how REMBO can find an acceptable subspace.
However, the straight lines are an indication for the hyperparameters to be no well chosen.
I do not allow hyperparameter optimization, for the reason such that the results may be comparable to the other algorithm's results.
As one can see, Camelback3D also provides a difficult for REMBO, although REMBO does find an acceptable subspace projection.

\begin{figure}[H]
\center
    \begin{subfigure}[b]{0.30\textwidth}
        \includegraphics[width=\textwidth]{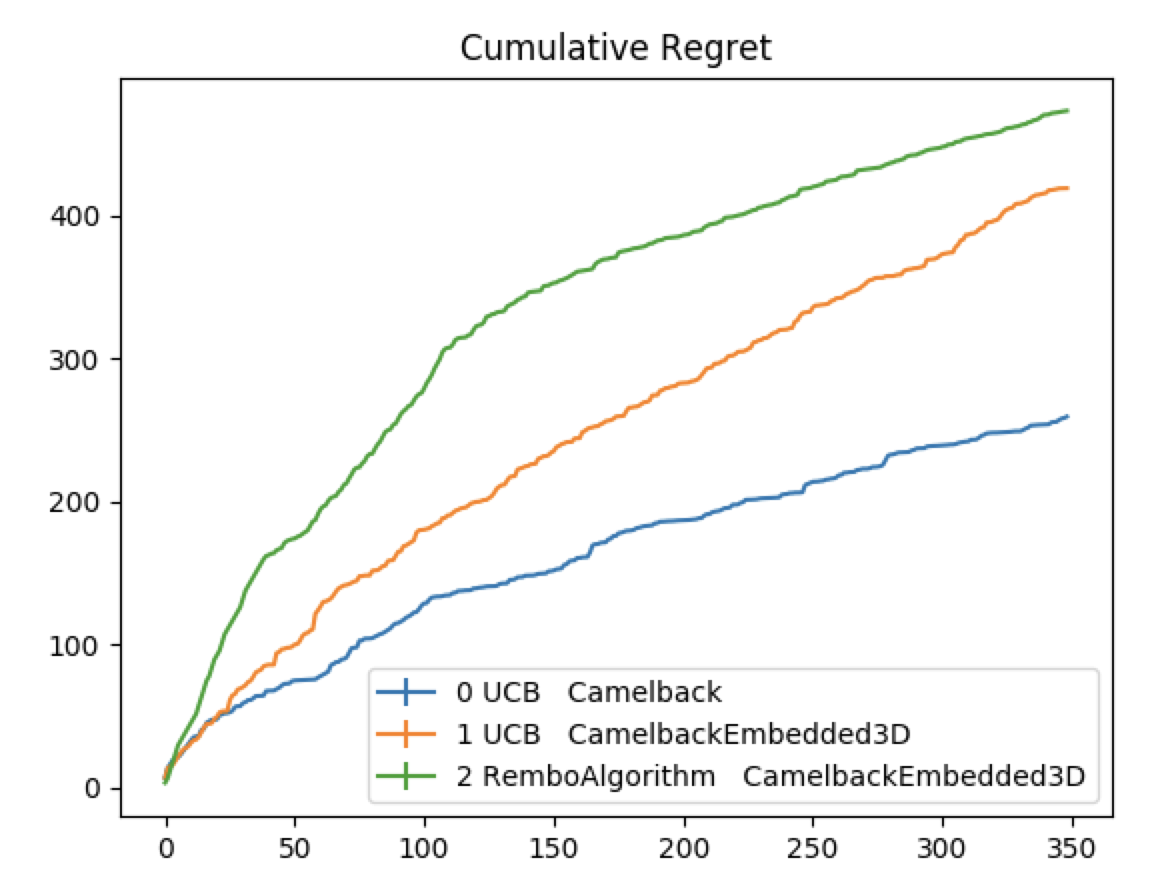}
        \label{fig:gull}
         \caption{Run 1}
    \end{subfigure}
        \begin{subfigure}[b]{0.30\textwidth}
        \includegraphics[width=\textwidth]{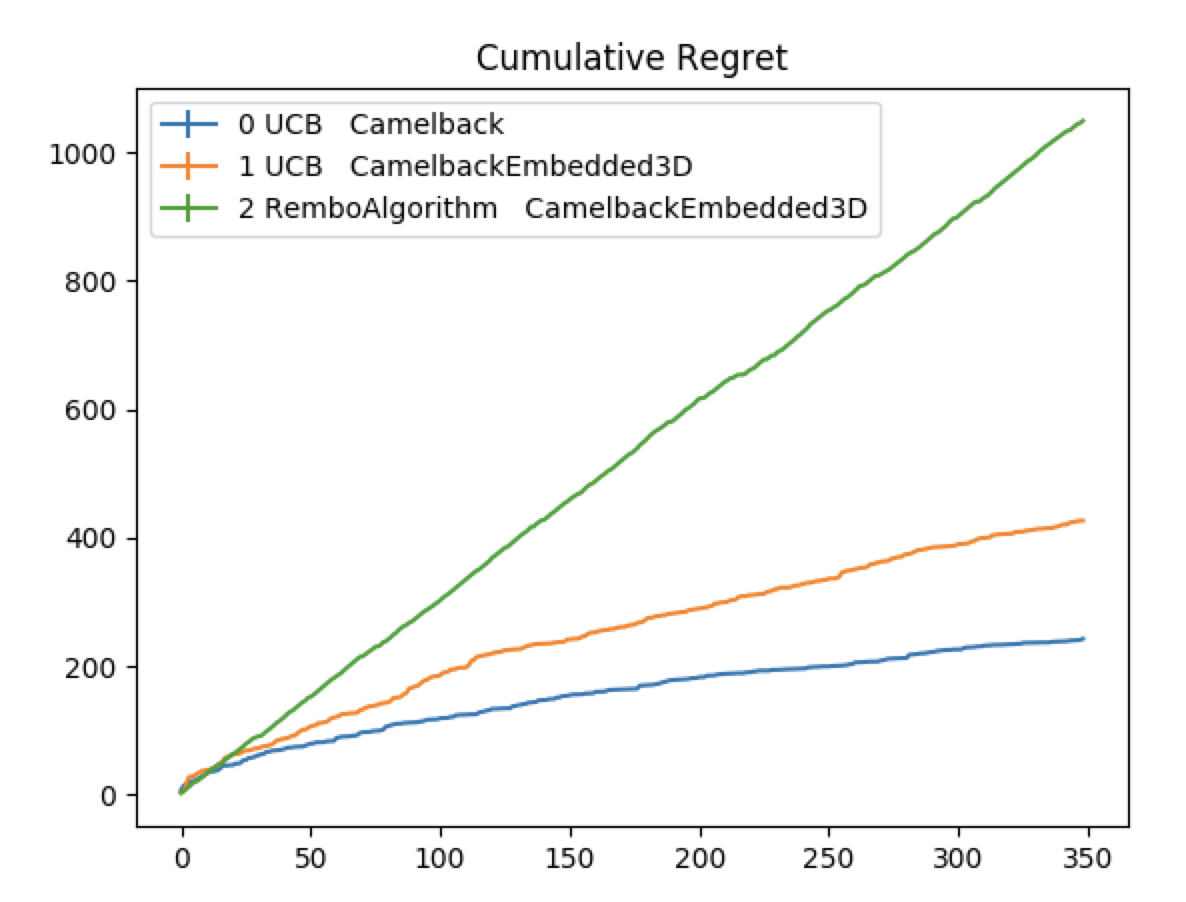}
        \label{fig:gull}
        \caption{Run 2}
    \end{subfigure}
    \begin{subfigure}[b]{0.30\textwidth}
        \includegraphics[width=\textwidth]{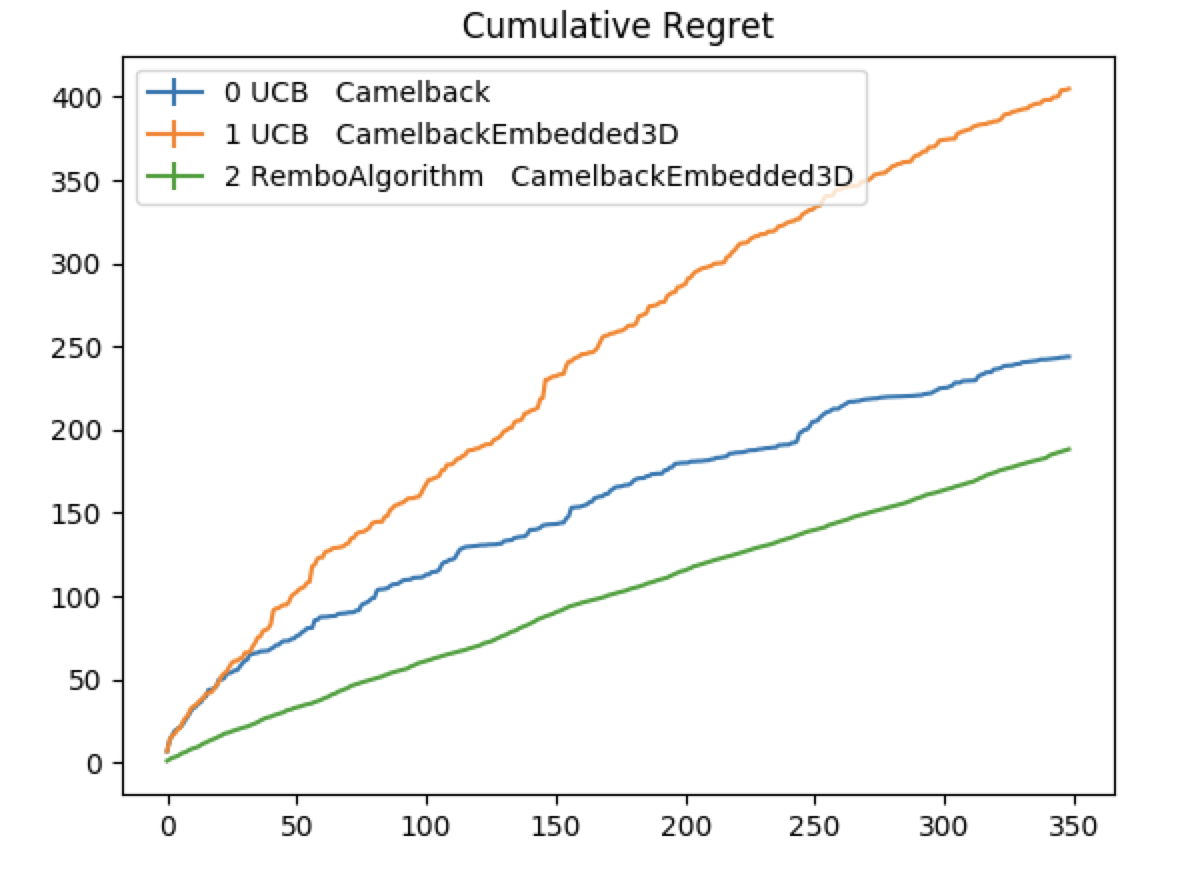}
        \label{fig:gull}
               \caption{Run 3}
    \end{subfigure}
        \caption{UCB using REMBO on a 2D Camelback embedded in a 3D space.
    }\label{fig:animals}
\end{figure}

\paragraph{Camelback5D}
The linearity of the UCB curves is still attained.
Similar to the result with Tripathy's method, REMBO can reduce the dimensionality of the subspace in almost any case.
The effectiveness of this reduction can be small though.
The viewer can recognize well that there is a high variance in the performance amongst runs.
This emphasizes how important the interleaved runs are, even though the number of data samples per projection is divided by the total number of projection (i.e., there is slower learning of the GP surface). However, dimensionality reduction is successful.

\begin{figure}[H]
\center
    \begin{subfigure}[b]{0.30\textwidth}
        \includegraphics[width=\textwidth]{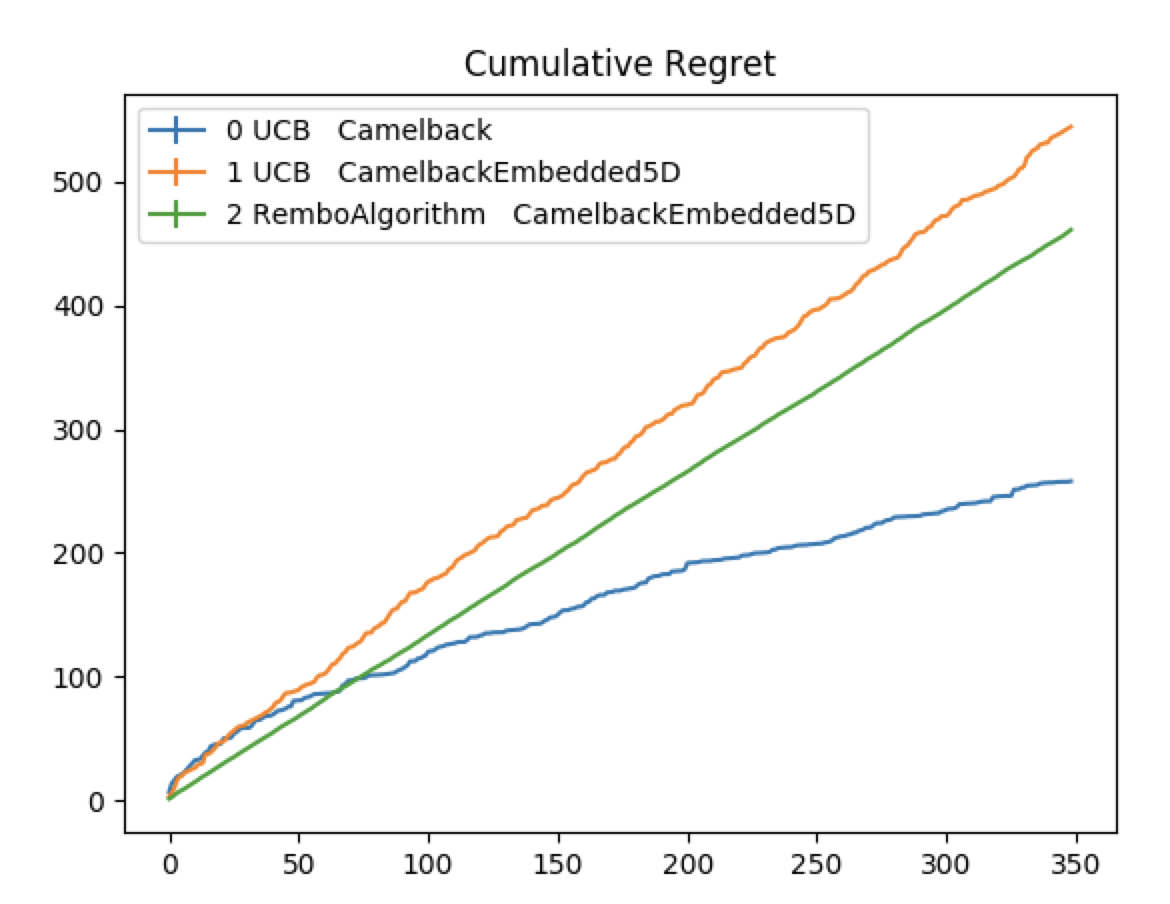}
        \label{fig:gull}
         \caption{Run 1}
    \end{subfigure}
        \begin{subfigure}[b]{0.30\textwidth}
        \includegraphics[width=\textwidth]{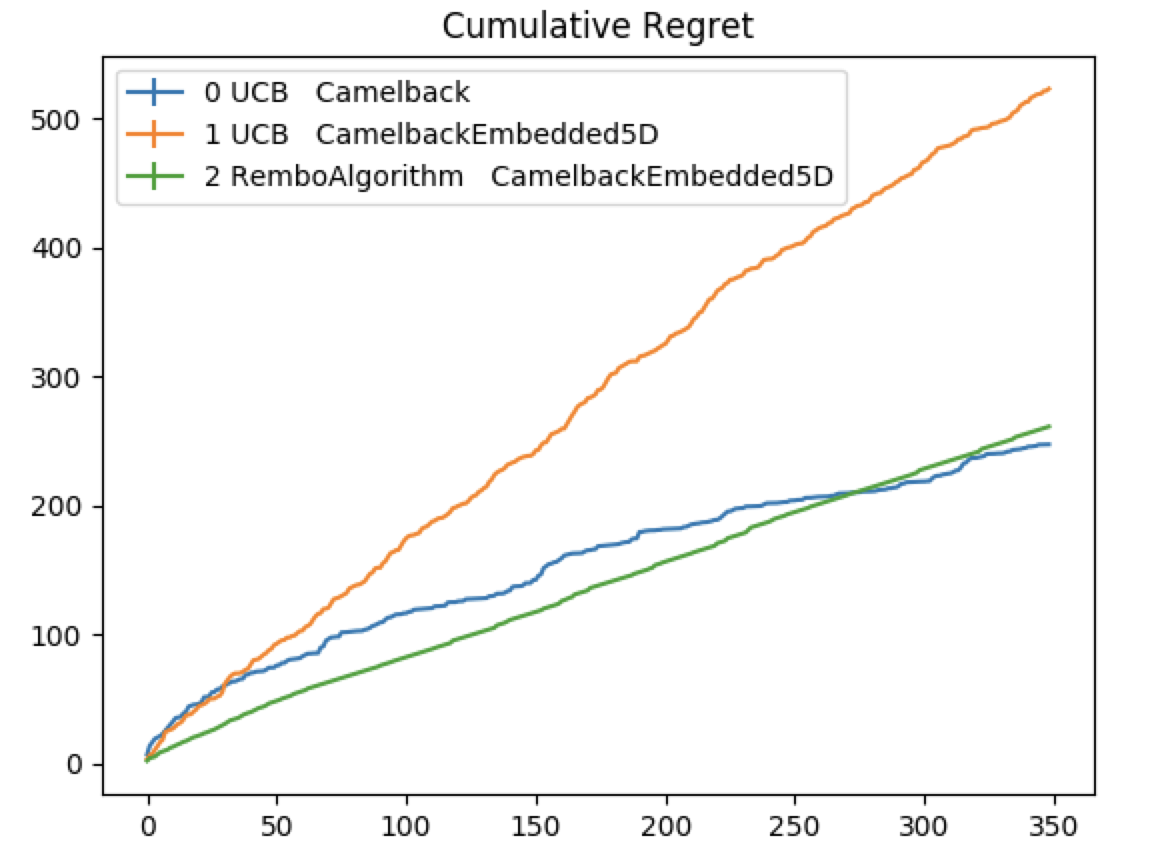}
        \label{fig:gull}
        \caption{Run 2}
    \end{subfigure}
    \begin{subfigure}[b]{0.30\textwidth}
        \includegraphics[width=\textwidth]{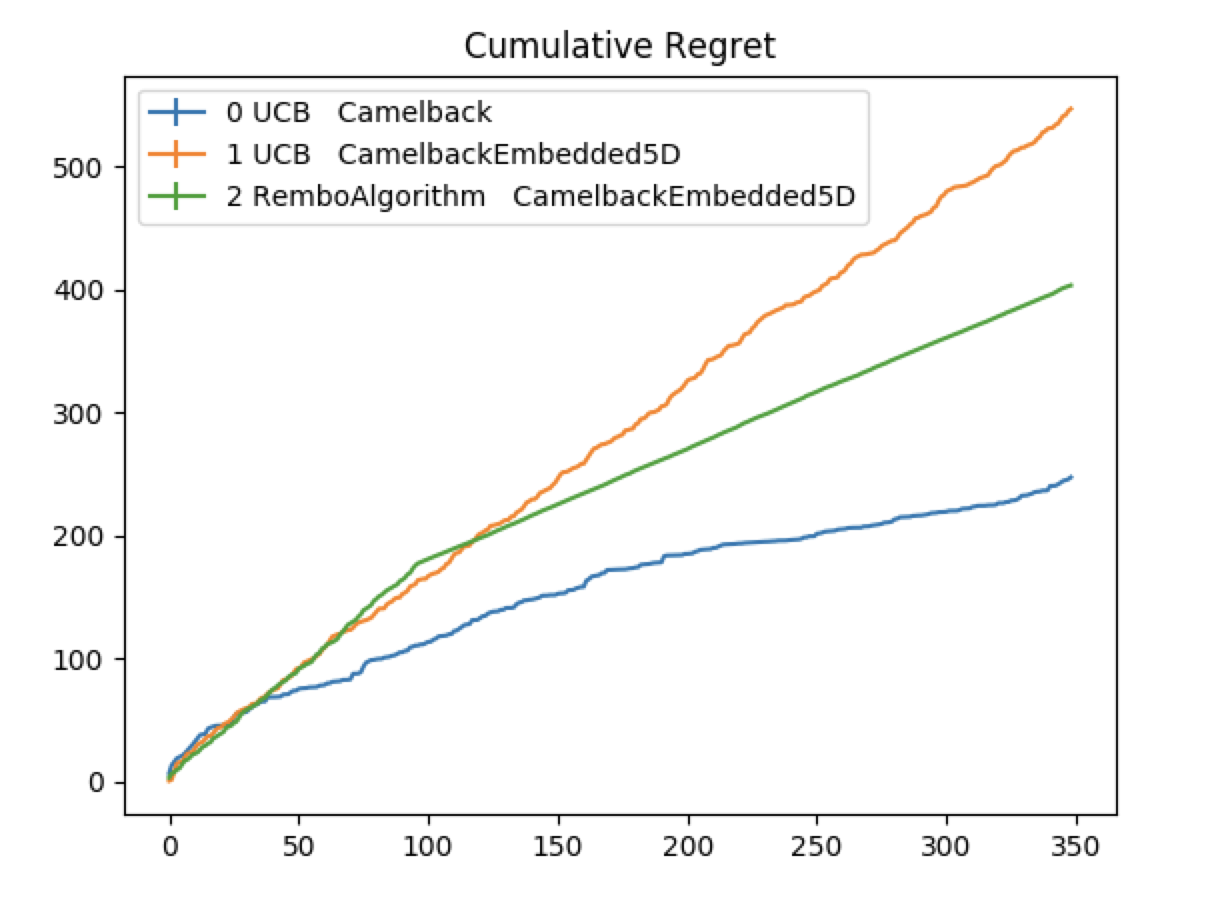}
        \label{fig:gull}
               \caption{Run 3}
    \end{subfigure}
        \caption{UCB using REMBO on a 2D Camelback embedded in a 5D space.
    }\label{fig:animals}
\end{figure}

\section{Qualitative evaluation}
It is interesting to see, that amongst the 1000 restarts that I generated some of the resulting matrices are close to the real projection matrix up to an absolute value of 0.01.
However, because the algorithm decides to choose the matrix with the highest likelihood, Tripathy's algorithm, in general, does not select these matrices but chooses a matrix that is not amongst the matrices that are very similar to the real matrices.

\subsection{Feature selection}
The goal of this task is to see if the revised active subspace identification algorithms can effectively apply feature selection.
For this task, I set up a function $ f $ that looks as follows:

\def\B{
\begin{bmatrix}
    (x - a_0)^2 \\
    (y - a_1)^2
\end{bmatrix}}

\begin{equation} \label{eq:FeatureExtension}
f \left( W \B \right) \approx g \left( x \right)
\end{equation} 

where $a_0, a_1$ are constants. \\

For this specific experiment, the function $f$ is chosen to be a one-dimensional parabola. 
As such, $W$ is chosen as a matrix on the Stiefel manifold with dimensions $\mathbf{R}^{d \times D}$.

Doing a feature extension over $x$ and $y$, we can get the following feature representation:

\def\PHI{
\begin{bmatrix}
    x_0^2 \\
    x_1^2 \\
    x_0 \\
    x_1 \\
    1
\end{bmatrix}}

\def\WtoPhi{
\begin{bmatrix}
    w_0 \\
    w_1 \\
    -2 w_0 a_0 \\
    -2 w_1 a_1 \\
    w_0 a_0^2 + w_1 a_1^2
\end{bmatrix}}

\begin{figure}[h]

\begin {minipage}{0.47\textwidth}
  \centering
  \begin{equation}
    \PHI
  \end{equation}
  \caption{Polynomial Kernel applied to vector $[x_0, x_1]$}
\end{minipage}
\hfill
\begin {minipage}{0.47\textwidth}
  \centering
  \begin{equation}
    \WtoPhi
  \end{equation}
  \caption{Corresponding weight matrix equivalent to \ref{eq:FeatureExtension} when applied on a parabola}
\end{minipage}

\end{figure}

To run experiments, I instantiate the "real" matrix, which should be found by the algorithm with the values $w_0 = 0.589$, $w_1 = 0.808$ (randomly sampled as a matrix on the Stiefel manifold), $a_0 = -0.1$, $a_1 = 0.1$ (chosen by me as coefficients). \\

I apply the algorithm 1. from \citep{Tripathy} to identify the active projection matrix.
The optimization algorithm has 50 samples to discover the hidden matrix, which it seemingly does up do a certain degree of accuracy.
Similar results are achieved for repeated tries.
The following figure shows the real matrix, and the matrix the algorithm has found.

\def\realW{
\begin{bmatrix}
    0.589 \\
    0.808 \\
    0.118 \\
    -0.162 \\
    0.823
\end{bmatrix}}

\def\okW1{
\begin{bmatrix}
    -0.355 \\
        -0.533 \\
        -0.908 \\
        0.099 \\
    -0.756 
\end{bmatrix}}

\begin{figure}[h] 
\begin {minipage}{0.47\textwidth}
  \centering
  \begin{equation} \label{fig:realMatrix}
    \realW
  \end{equation}
  \caption{Real matrix}
\end{minipage}
\hfill
\begin {minipage}{0.47\textwidth}
  \centering
  \begin{equation} \label{fig:foundMatrix}
    \okW1
  \end{equation}
  \caption{Matrix found by optimization algorithm}
\end{minipage}
\end{figure}

Although the element-wise difference between the two matrices \ref{fig:realMatrix} and \ref{fig:foundMatrix} is high (between $0.05$ and $0.15$, one entry is above $1$), one can see that the matrix recovery is successful in finding an approximate structure that resembles the original structure of the features.
One should observe that the found matrix is an approximate solution to the real matrix in the projection. I.e., the matrix found is close to the real matrix, but multiplied by $-1$. \\

Because in this case, I applied the feature selection algorithm on a vector-matrix (only one column), one can quantify the reconstruction of the real matrix through the found matrix by the normalized scalar product.
This quantity is a metric between $0$ and $1$, where $0$ means that both vectors are orthogonal, and $1$ means that both vectors overlap.

\begin{equation}
\text{overlap}(u, v) = \frac{| \langle u, v \rangle |}{\langle u, u \rangle}
\end{equation}
where $u$ is the real vector, and $v$ is the found vector.

Inserting the actual values into the field, we get $0.79$, which is a good value for the feature vector found, and the trained number of data points which is 50. \\
 
 This experiment shows that algorithm 1. from \citep{Tripathy} successfully allows a viable option to other feature selection algorithms, by providing a measure, where the optimal linear projection is found. 
 However, one must notice that other feature selection algorithms (such as SVM \citep{SVMFeature}), are more efficient, and will provide better results with a higher probability if applied on a similar kernel. \\
 
 One observation I made was the increase in the log-likelihood of the data w.r.t. the projection matrix did not correlate with the decrease in the angle between the real vs. the found projection matrix.
 Also, most often, the angle was at around 40 degrees, which means that only slight improvements over an entirely random embedding were made.

\subsection{Subspace identification}
One of the main reasons to use our method is because we allow for subspace identification.
We have the following functions:

\begin{enumerate}
\item 1D Parabola embedded in a 2D space
\item 2D Camelback embedded in a 5D space
\item 2D Sinusoidal and Exponential function embedded in a 5D space (see Appendix)
\end{enumerate}

To be able to visualize the points, I proceed with the following procedure:

I generate testing points (points to be visualized) within the 2D-space in a uniform grid.
I then project these testing points to the dimension of the original function ($2d$ for parabola, else $5d$).
I then let each algorithm learn and predict the projection matrix, and GP mean predictions.
If because the transformation from $2D$ space to $5D$ space and GP mean prediction is each bijective, we can visualize the $2D$ points with the GP mean prediction right away.
As such, the dimension of the embedding learned does not have an impact on the visualization.

In the following figures, blue point shows the sampled real function value.
Orange points show the sampled mean prediction of the trained GP.
The GPs were each trained on 100 data points. 
The points shown below were not used for training at any point, as these are included in the test set.

\begin{figure}[H]
    \centering
    \begin{subfigure}[b]{0.30\textwidth}
        \includegraphics[width=\textwidth]{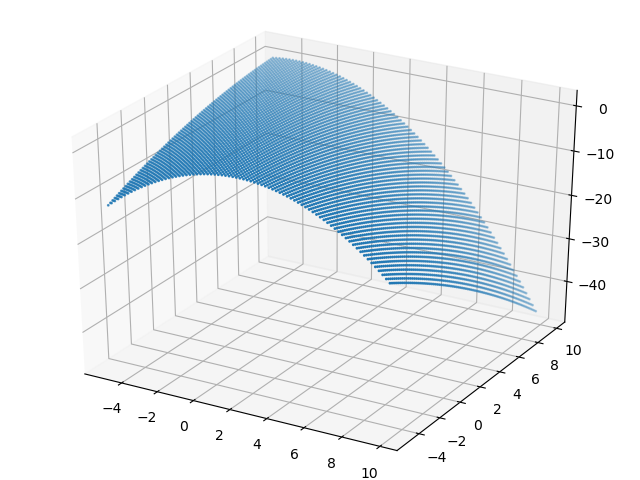}
        \caption{Parabola Original}
        \label{fig:gull}
    \end{subfigure}
    \begin{subfigure}[b]{0.30\textwidth}
        \includegraphics[width=\textwidth]{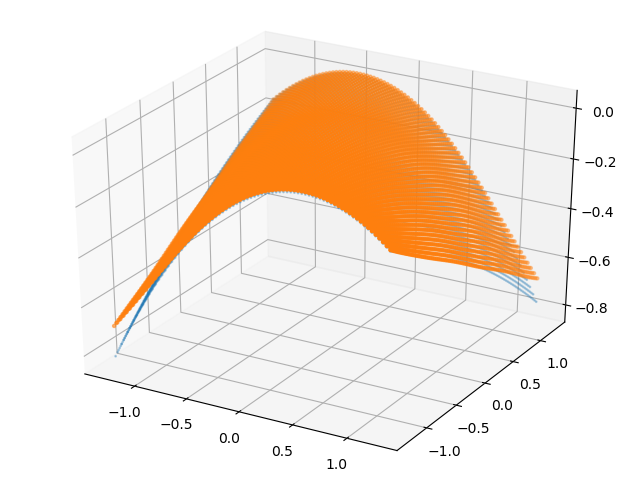}
        \caption{Parabolanusoidal Boring}
        \label{fig:tiger}
    \end{subfigure}
    \vskip\baselineskip
    \begin{subfigure}[b]{0.30\textwidth}
        \includegraphics[width=\textwidth]{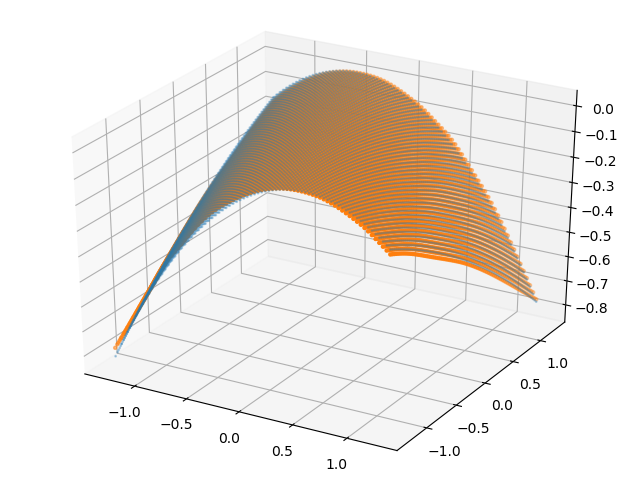}
        \caption{Parabola Tripathy}
        \label{fig:mouse}
    \end{subfigure}
    \begin{subfigure}[b]{0.30\textwidth}
        \includegraphics[width=\textwidth]{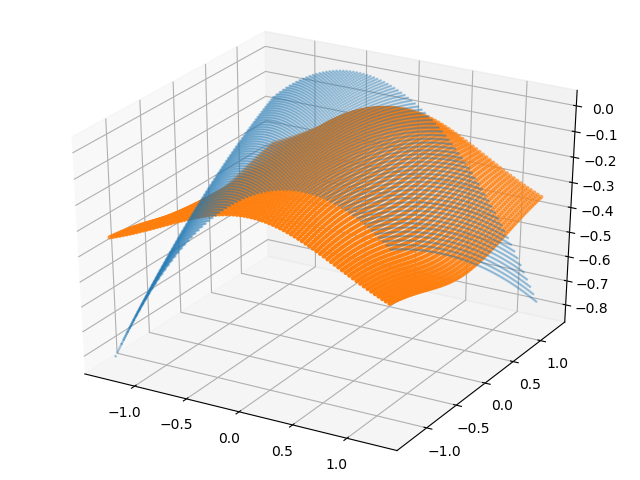}
        \caption{Parabola Rembo}
        \label{fig:mouse}
    \end{subfigure}
    \caption{Top-Left: The 1D Parabola which is embedded in a 2D space.}\label{fig:animals}
\end{figure}

I set the number of restarts to $14$ and number of randomly sampled data points to 100.
Notice that the Tripathy approximation is slightly more accurate than the BORING approximation. 
This is because one of Tripathy's initial starting points were selected better, such that algorithm 3 ran many times before the relative loss terminated the algorithm.
The active subspace projection matrix is of size $\mathbf{R}^{1 \times 2}$

\begin{figure}[H]
    \centering
    \begin{subfigure}[b]{0.25\textwidth}
        \includegraphics[width=\textwidth]{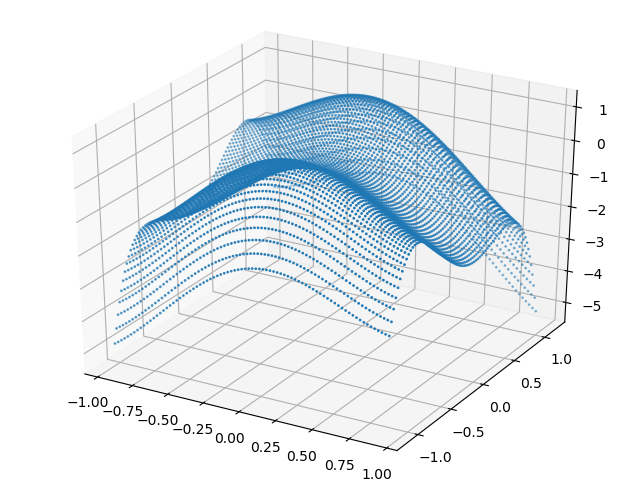}
        \caption{Camelback Original}
        \label{fig:gull}
    \end{subfigure}
    ~ 
    \begin{subfigure}[b]{0.25\textwidth}
        \includegraphics[width=\textwidth]{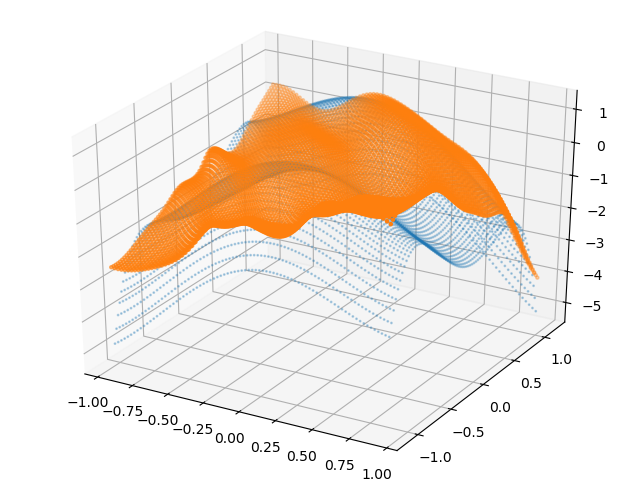}
        \caption{Camelback Boring}
        \label{fig:tiger}
    \end{subfigure}
        \vskip\baselineskip
    \begin{subfigure}[b]{0.25\textwidth}
        \includegraphics[width=\textwidth]{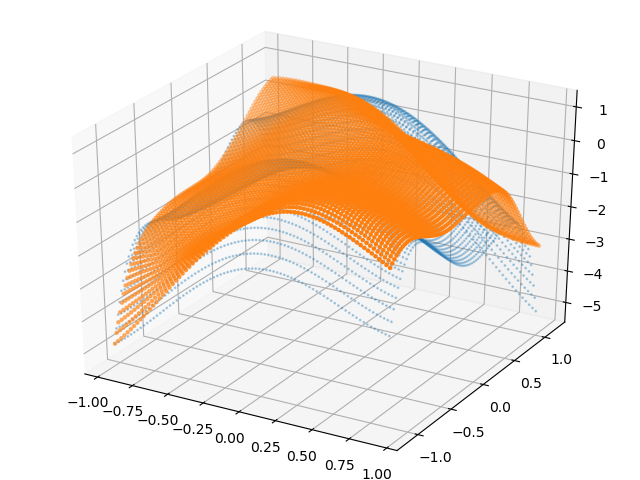}
        \caption{Camelback Tripathy}
        \label{fig:mouse}
    \end{subfigure}
        ~ 
    \begin{subfigure}[b]{0.25\textwidth}
        \includegraphics[width=\textwidth]{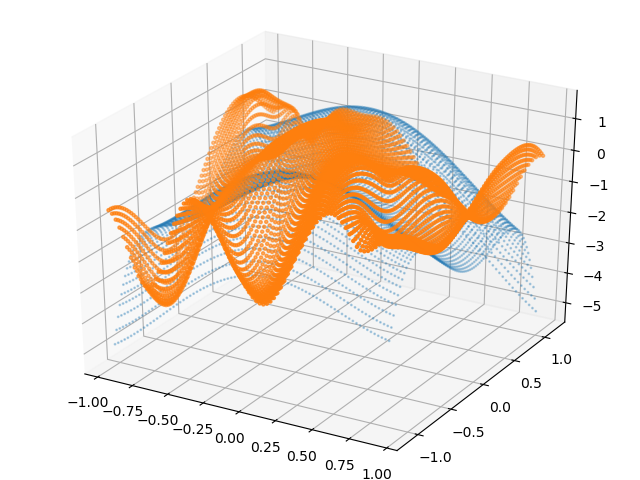}
        \caption{Camelback Rembo}
        \label{fig:mouse}
    \end{subfigure}
    \caption{Top-Left: The 2D Camelback Function which is embedded in a 5D space.}\label{fig:animals}
\end{figure}

I set the number of restarts to $28$ and number of randomly sampled data points to 100.
The active subspace projection matrix is of size $\mathbf{R}^{2 \times 5}$, as this is a function that lives in a 2D space, and has two strong principal components.
Notice that Tripathy and BORING use the same algorithm, as the visualization does not allow to add a third axis. 
In other words, BORING does not add any additional orthogonal vector to the model.
As such, it does not add any additional kernels to the model as well and is equivalent to Tripathy.

\chapter{Conclusion}

\ifpdf
    \graphicspath{{Chapter7/Figs/Raster/}{Chapter7/Figs/PDF/}{Chapter7/Figs/}}
\else
    \graphicspath{{Chapter7/Figs/Vector/}{Chapter7/Figs/}}
\fi

\section{Future work}

Future work could incorporate the synthesis of different methods, including additive GPs and Tripathy's method.
Although Tripathy's method is unbeaten in identifying the active subspace dimension, heuristics could be easily implemented to speed up the calculation time.
To avoid the (small) possibility of identifying a bad subspace, one could make use of the idea of interleaved runs as used in REMBO on Tripathy's method to marginalize this probability even further.
One of the most critical aspects is hyperparameter tuning for the GP models itself.
These can make or break the identification of the subspace.
Choosing bad hyperparameters does not allow us to compare different methods effectively.
In the future, it would be beneficial to address this issue to a stronger extent.
Recalculating the search space for each new point may be too time-consuming, with which REMBO would still be considered state of the art regarding Bayesian Black Box Optimization.



\begin{spacing}{0.9}


\bibliographystyle{apalike}
\cleardoublepage
\bibliography{thesis} 



\end{spacing}


\begin{appendices} 

\ifpdf
    \graphicspath{{Appendix1/Figs/Raster/}{Appendix1/Figs/PDF/}{Appendix1/Figs/}}
\else
    \graphicspath{{Appendix1/Figs/Vector/}{Appendix1/Figs/}}
\fi

\chapter{Appendix A}

\section{Benchmarking functions}

The following is a list of synthetic functions and real datasets presented in previous works, where the goal is to evaluate Bayesian optimization algorithms. 

\begin{enumerate}
\item \citep{Garnett2013} Synthetic in-model data matching the proposed model, with $d=2, 3$, and $D=10, 20$.
\item \citep{Garnett2013} (Synthetic) Braning function, $d=2$, hidden in a higher dimensional space $D=10, 20$.
\item \citep{Garnett2013} Temperature data $D=106$ and $d=2$.
\item \citep{Garnett2013} Communities and Crime dataset $d=2$, and $D=96$. 
\item \citep{Garnett2013} Relative location of CT slices on axial axis with $d = 2$ and $D=318$. 
\item \citep{Djolonga2013} (Synthetic) Random GP samples from 2-dimensional Matern-Kernel-output, embedded within 100 dimensions
\item \citep{Djolonga2013} Gabor Filters: Determine visual stimuli that maximally excite some neurons which reacts to edges in the image.
We have $f(x) = \exp( -( \theta^T x - 1 )^2 )$. $\theta$ is of size 17x17, and the set of admissible signals is $d$.
\item \citep{Wang2013} (Synthethic) $d=2$ and $D=1*10^9$.
\item \citep{Wang2013} $D=47$ where each dimension is a parameter of a mixed integer linear programming solver.
\item \citep{Wang2013} $D=14$ with $d$ for a random forest body part classifier.
\item \citep{Tripathy}  (Synthetic) Use $d=1,10$ and $D=10$.
\item \citep{Tripathy} (Half-synthetic) Stochastic elliptic partial differential equation, where $D=100$, and an assume value for $d$ of $1$ or $2$.
\item \citep{Tripathy} Granular crystals $X \in \mathbb{R}^{1000 \times 2n_p +1}$, and $y \in \mathbb{R}^{1000}$.
\item \citep{Gardner2017} (Synthetic) Styblinski–Tang function where $D$ is freely choosable.
\item \citep{Gardner2017} (Synthetic) Michalewicz function where $D$ is freely choosable.
\item \citep{Gardner2017} (Simulated) NASA cosmological constant data where $D=9$.
\item \citep{Gardner2017} Simple matrix completion with $D=3$.
\item \citep{Rana2017} (Synthetic) Hertmann6d in [0, 1].
\item \citep{Rana2017} (Synthetic) Unnormalized Gaussian PDF with a maximum of 1 in $[-1, 1]^d$ for $D=20$ and $[-0.5, 0.5]^d$ for $D=50$
\item \citep{Rana2017} (Synthetic) Generalized Rosenbrock function $[-5, 10]^d$
\item \citep{Rana2017} Training cascade classifiers, with $D=10$ per group.
\item \citep{Rana2017} Optimizing alloys $D=13$. 
\item \citep{Li2018} (Synthetic) Gaussian mixture function
\item \citep{Li2018} (Synthetic) Schwefel's 1.2 function.
\item \citep{OptimizationTestFunctions} A list of optimiztation test functions can be found here.
\item \citep{Jamil2013} A more comprehensive list of general functions can be found here.
\end{enumerate}

\section{The log-likelihood function as a function of $\tau$}
In \ref{Eq:LogLikelihoodF} I present the formula for the log-likelihood function which is a function of $W$.
In \ref{Eq:TauFunction} I show how modifying a scalar parameter $\tau$ can modify this loss.
In a combined formula, this looks as follows:

\begin{equation}
G: \tau \rightarrow F( \gamma(\tau; W_{ \text{fixed} }) )
\end{equation}
where we assume that $W_{ \text{fixed} }$ is fixed. 
To get an intuition for how smooth this function $G$ is, I randomly sample $W$ and visualize the change of $G(\tau)$ for different values of $\tau$.\\

\begin{figure}[H]
    \centering
    \begin{subfigure}[b]{0.40\textwidth}
        \includegraphics[width=\textwidth]{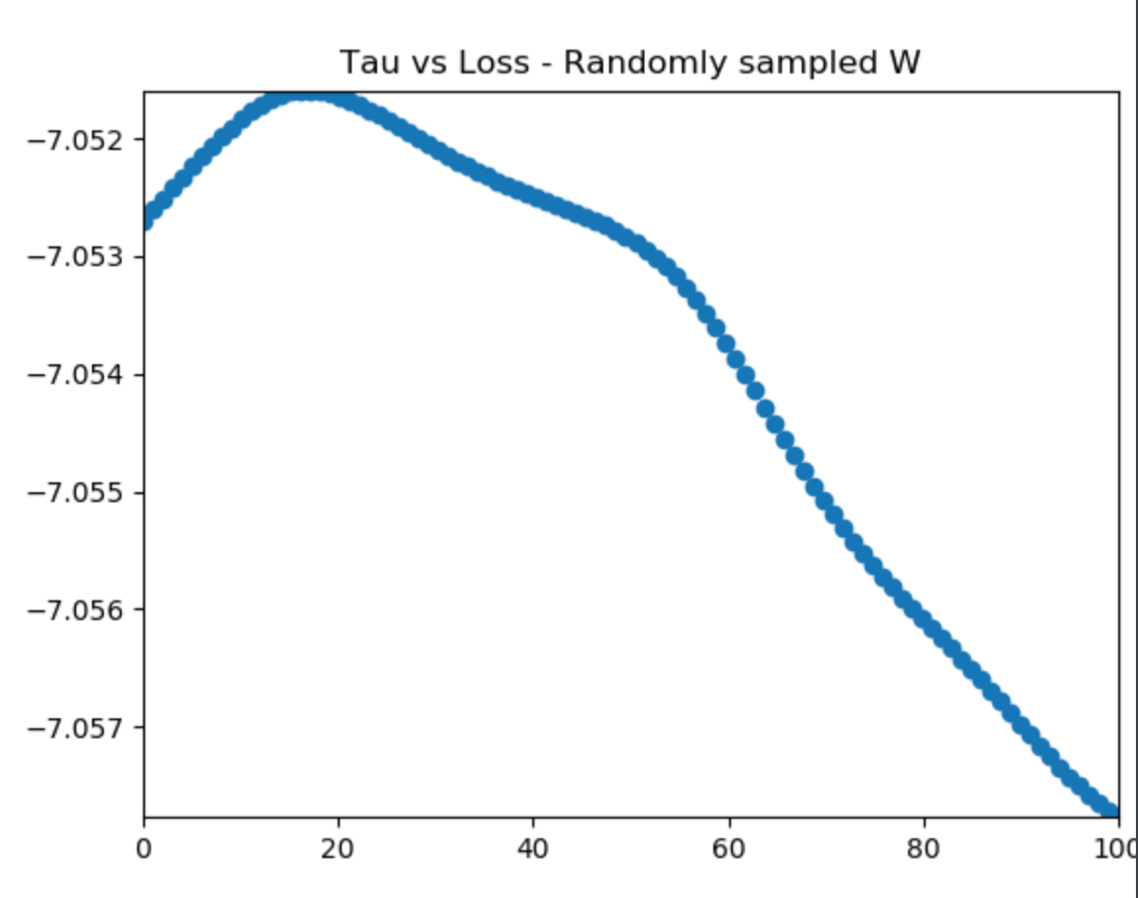}
        \label{fig:gull}
        \caption{W is an orthogonal matrix}
    \end{subfigure}
    \quad
    \begin{subfigure}[b]{0.40\textwidth}
        \includegraphics[width=\textwidth]{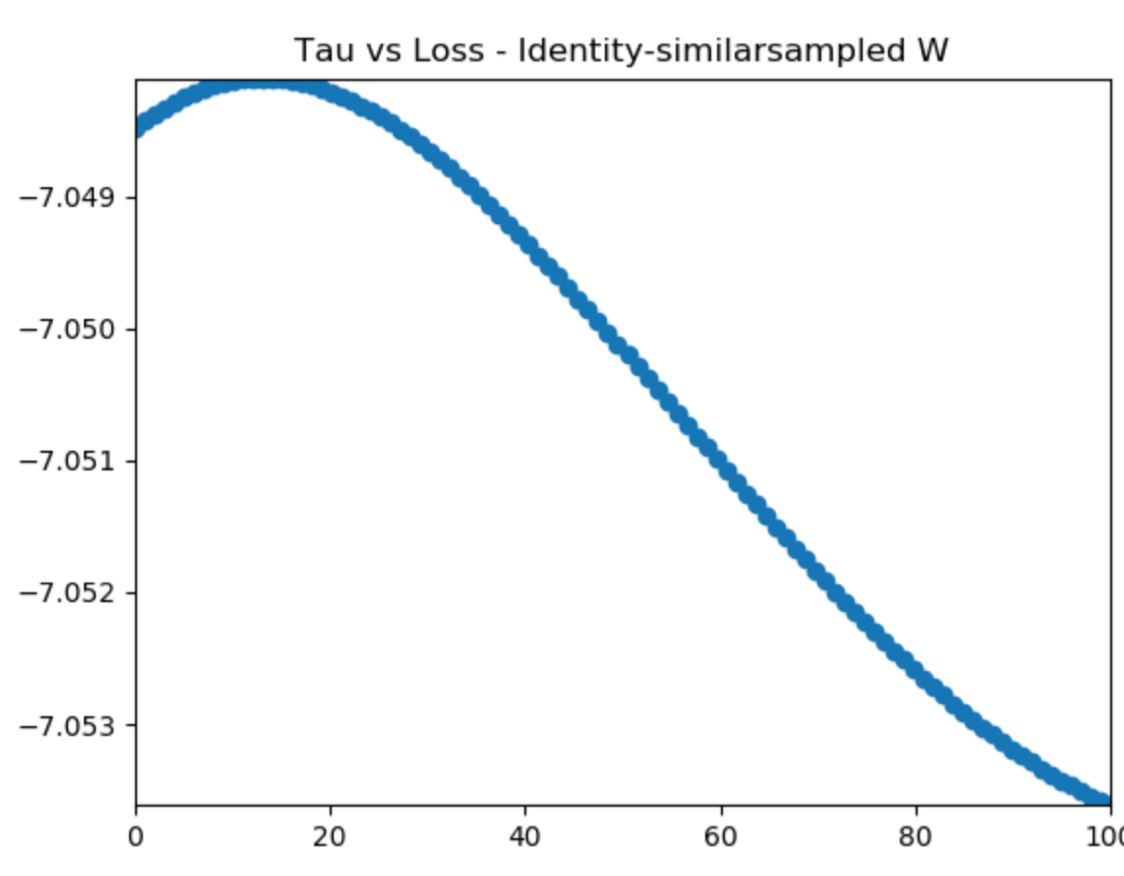}
        \label{fig:tiger}
        \caption{W is taken to be the identity matrix}
    \end{subfigure}   
           \caption{Perturbing $\tau$ and getting a log-likelihood value through $G(\tau)$.
           The function at hand is a Parabola embedded in a 2D space.
           The x-axis refers to the individually sampled $\tau$ values (0 is $\tau = 0$, and 100 is $\tau = 1$ and every value in between is linearly interpolated).
           The y-axis refers to the log-likelihood.
           One can see that the function is smooth on this scale, and that the meaximum is achieved between $0 \leq \tau \leq 0.3 $. }
\end{figure}

\section{Additional Visualization}

The following is a visualization of the sinusoidal function which we use in first part of the quantitave evaluation (i.e. UCB runs):

\begin{figure}[H]
\centering
\begin{subfigure}[b]{0.4\textwidth}
	\includegraphics[width=\textwidth]{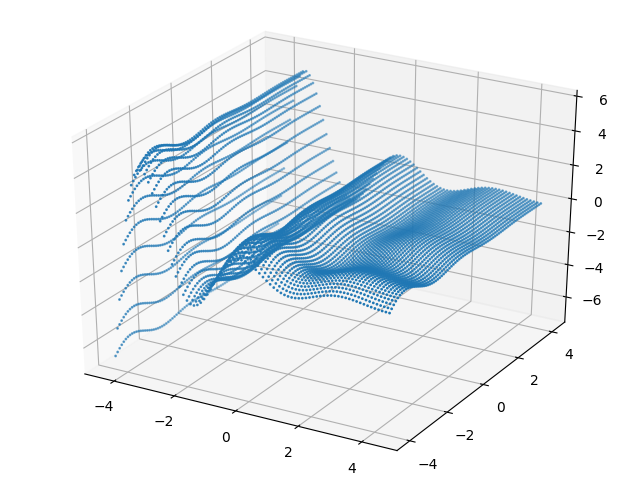}
	\caption{Sinusoidal used for UCB analysis}
\end{subfigure}
\end{figure}

The above visualization is in contrast to the function which I use in the second part of the quantitative analysis (measuring development of angle loss and log-likelihood), which I present below.
The difference between the two functions is the embeddings, which naturally capture a different range.
However, the embedding is different, and was thus easy to find by Tripathy's method.
This is the same function whose angle Tripathy's algorithm tries to find in chapter 6.

\begin{figure}[H]
    \centering
    \begin{subfigure}[b]{0.25\textwidth}
        \includegraphics[width=\textwidth]{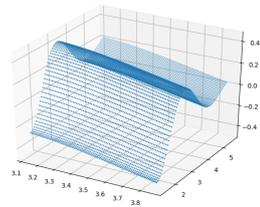}
        \caption{Sinusoidal Original}
        \label{fig:gull}
    \end{subfigure}
    ~ 
    \begin{subfigure}[b]{0.25\textwidth}
        \includegraphics[width=\textwidth]{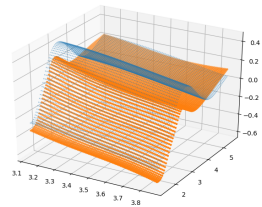}
        \caption{Sinusoidal Boring}
        \label{fig:tiger}
    \end{subfigure}
        \vskip\baselineskip
    \begin{subfigure}[b]{0.25\textwidth}
        \includegraphics[width=\textwidth]{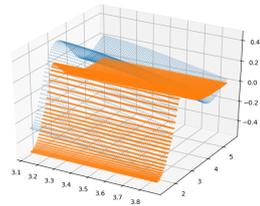}
        \caption{Sinusoidal Tripathy}
        \label{fig:mouse}
    \end{subfigure}
        ~ 
    \begin{subfigure}[b]{0.25\textwidth}
        \includegraphics[width=\textwidth]{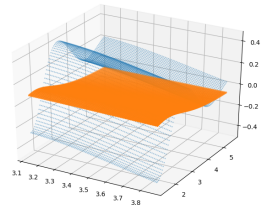}
        \caption{Sinusoidal Rembo}
        \label{fig:mouse}
    \end{subfigure}
    \caption{Top-Left: The 2D Sinusoidal-Exponential Function which is embedded in a 5D space.}\label{fig:animals}
\end{figure}

I set the number of restarts to $28$ and number of randomly sampled data points to 100.
The active subspace projection matrix is of size $\mathbf{R}^{1\times 5}$, as this is a function that exhibits a strong principal component, but that still attains small perturbations among a different dimension.
One can see very well here, that BORING can take into account the small perturbations, at a considerably lower cost than Tripathy would be able to.

\end{appendices}

\printthesisindex 

\end{document}